\documentclass[10pt,twocolumn,letterpaper]{article}

\usepackage{iccv}
\usepackage{times}
\usepackage{epsfig}
\usepackage{graphicx}
\usepackage{amsmath}
\usepackage{amssymb}
\usepackage{csquotes}
\usepackage{float}

\usepackage{times}
\usepackage{epsfig}
\usepackage{graphicx}
\usepackage{amsmath,amssymb,amsthm}
\usepackage{bm,xspace}
\usepackage{verbatim}
\usepackage{multirow}
\usepackage{wrapfig}
\usepackage{balance}

\def\nn{\mathbf{n}}

\def\pp{\mathbf{p}}
\def\qq{\mathbf{q}}

\def\xx{\mathbf{x}}

\def\zz{\mathbf{z}}

\def\TT{\mathbf{T}}

\def\C{\mathcal{C}}

\def\K{\mathcal{K}}
\def\cL{\mathcal{L}}

\def\N{\mathcal{N}}
\def\OC{\mathcal{O}}
\def\P{\mathcal{P}}

\def\S{\mathcal{S}}
\def\T{\mathcal{T}}

\def\Re{\mathbb{R}}

\def\btheta{{\bm\theta}}

\def\eps{\varepsilon}

\def\rot#1{\rotatebox{90}{#1}}

\newcommand\mypara[1]{\vspace{1mm}\noindent\textbf{#1}}

\usepackage{tabularx}
\newcommand{\PBS}[1]{\let\temp=\\#1\let\\=\temp}
\newcommand{\RBS}{\let\\=\tabularnewline}

\DeclareMathOperator{\pre}{precision}

\DeclareMathSymbol{@}{\mathord}{letters}{"3B}

\usepackage[pagebackref=true,breaklinks=true,letterpaper=true,colorlinks,bookmarks=false]{hyperref}

\iccvfinalcopy 

\def\iccvPaperID{254} 

\ificcvfinal\fi
\begin{document}

\title{Learning Compact Geometric Features}

\author{Marc Khoury\\
UC Berkeley
\and
Qian-Yi Zhou\\
Intel Labs
\and
Vladlen Koltun\\
Intel Labs
}

\maketitle

\begin{abstract}
We present an approach to learning features that represent the local geometry around a point in an unstructured point cloud. Such features play a central role in geometric registration, which supports diverse applications in robotics and 3D vision. Current state-of-the-art local features for unstructured point clouds have been manually crafted and none combines the desirable properties of precision, compactness, and robustness. We show that features with these properties can be learned from data, by optimizing deep networks that map high-dimensional histograms into low-dimensional Euclidean spaces. The presented approach yields a family of features, parameterized by dimension, that are both more compact and more accurate than existing descriptors.
\end{abstract}


\section{Introduction}
\label{sec:introduction}

Local geometric descriptors represent the local geometry around a point in a point cloud. They play a central role in geometric registration, which supports diverse applications in robotics and 3D vision \cite{Holz2015} and underpins modern 3D reconstruction pipelines~\cite{Zhou2016}. To enable accurate and efficient registration, the descriptor must possess a number of properties~\cite{Guo2016}. First, it should map the local geometry to a vector in a Euclidean space $\Re^n$; such Euclidean representations support efficient geometric search structures and nearest-neighbor queries. Second, the descriptor should be discriminative: nearest neighbors in feature space should correspond to points with genuinely similar local neighborhoods. Third, the representation should be compact, with a small dimensionality $n$: this supports fast spatial search. Finally, the representation should be robust to artifacts that are commonly encountered in real data, such as noise and missing regions.

\begin{figure}[t]
\centering
    \includegraphics[width = 1\linewidth]{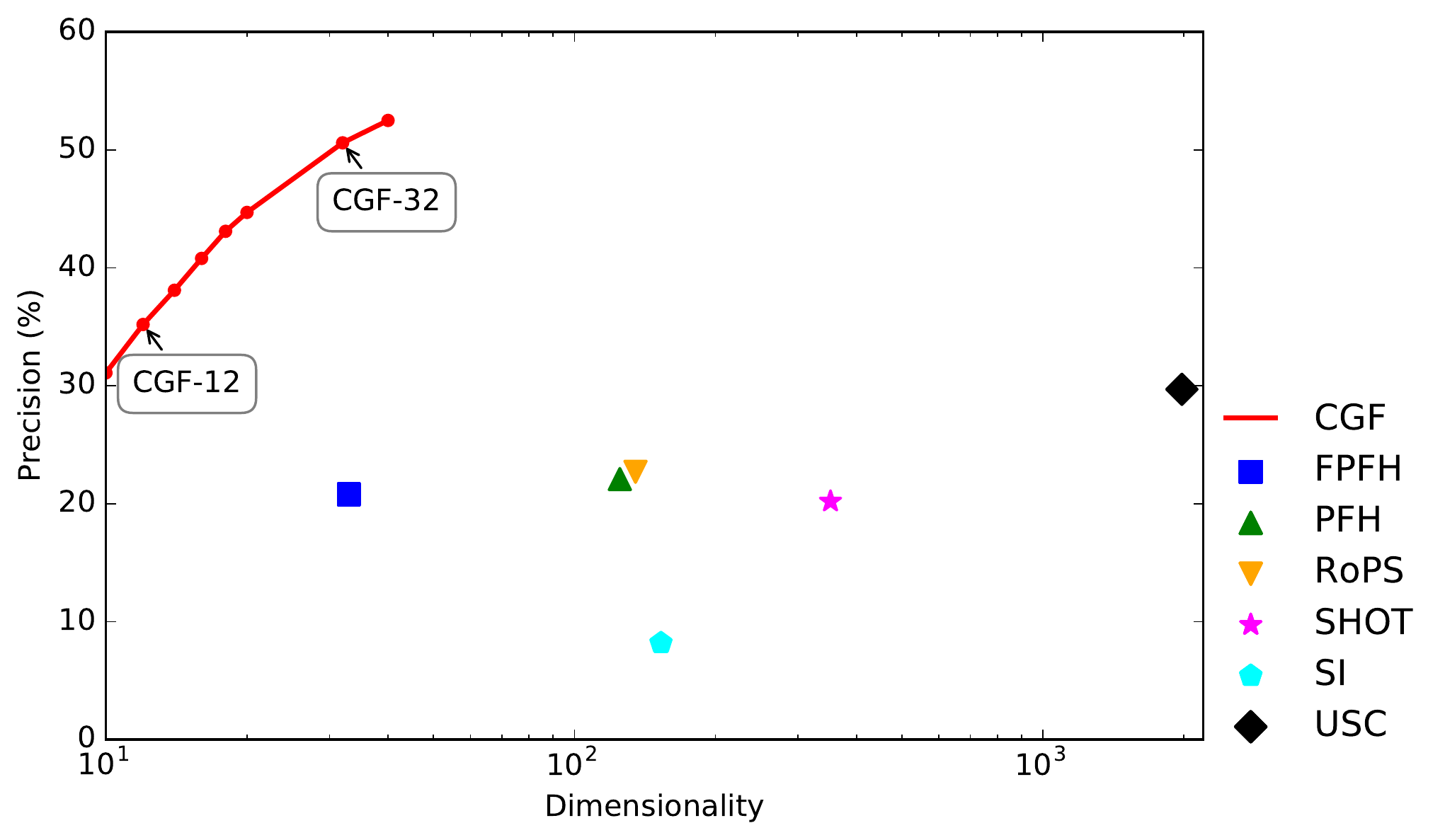}
\caption{Our approach yields a family of Compact Geometric Features (CGF), parameterized by dimension. This figure illustrates the performance of CGF on the SceneNN test set. Our features are both more compact and more precise than the baselines. The horizontal axis (dimensionality) is on a logarithmic scale.}
\label{fig:tradeoff}
\end{figure}

The design of local geometric descriptors has been the subject of intensive study for the past two decades. Many hand-crafted descriptors have been designed and evaluated \cite{JohnsonHebert1999,Frome2004,Rusu2009,Salti2014}. Nevertheless, no existing descriptor jointly satisfies the desiderata of high discriminative ability, compactness, and robustness~\cite{Guo2016}. Part of the challenge is the difficulty of optimizing the parameters of a high-dimensional feature representation by hand.

In this paper, we present an approach to learning local geometric features from data. Our descriptor applies directly to unstructured point clouds and does not require a clean and consistent surface parameterization~\cite{Boscaini2016}, a volumetric representation~\cite{Zeng2016}, or the synthesis of auxiliary depth images~\cite{Wei2016}. Our features support nearest-neighbor queries in a Euclidean space, which allows establishing dense correspondences across point sets in near-linear time, in contrast to the quadratic complexity required by pairwise matching networks. We thus obtain the first learned geometric feature that can serve as a drop-in replacement for state-of-the-art hand-crafted features in existing pipelines~\cite{Holz2015,Zhou2016}.

We show that the presented approach yields descriptors that are both more discriminative and more compact than state-of-the-art hand-crafted features.
An illustration is provided in Figure~\ref{fig:tradeoff}. Experiments demonstrate that our Compact Geometric Features (CGF) yield more accurate matches at lower query times. When CGF is used on the standard Redwood benchmark for geometric registration, with no training or fine-tuning on that dataset, it yields the highest recall reported on the benchmark to date.

\section{Background}
\label{sec:background}
The development of geometric descriptors for rigid alignment of unstructured point clouds dates back to the 90s. Classic descriptors include Spin Images~\cite{JohnsonHebert1999} and 3D Shape Context~\cite{Frome2004}. More recent work introduced Point Feature Histograms (PFH) \cite{Rusu2008-IROS}, Fast Point Feature Histograms (FPFH) \cite{Rusu2009}, Signature of Histogram Orientations (SHOT) \cite{Salti2014}, and Unique Shape Contexts (USC) \cite{Tombari2010b}. A comprehensive evaluation of existing local geometric descriptors is reported by Guo et al.~\cite{Guo2016}.

Significant work has also been conducted on descriptors for nonrigid registration of deformable surfaces~\cite{Aubry2011,Sun2009}. These descriptors tend to make stronger assumptions, such as the existence of a reasonably clean meshed surface, and are designed to be invariant to isometric deformations. In contrast, rigid registration requires sensitivity to isometric deformations -- the opposite of invariance. And applications in robotics require handling noisy unstructured point sets. Our work is devoted to rigid registration of unstructured point clouds.

A number of recent works applied learning to the problem of matching corresponding points based on local geometry. Wei et al.~\cite{Wei2016} describe an approach that matches points on human body scans and operates on ensembles of depth images. Boscani et al.~\cite{Boscaini2016} extend convolutional networks to Riemannian manifolds and apply them to establish correspondences across compatible manifolds. The contemporaneous work of Zeng et al.~\cite{Zeng2016} uses volumetric signed distance fields and develops learned descriptors that use such volumetric representations as input. Cosmo et al.~\cite{Cosmo2016} learn descriptors for isometry-invariant nonrigid matching. In contrast, our work is devoted to learning compact descriptors for local geometry in unstructured point clouds, which can be used as highly efficient drop-in replacements for prior such descriptors in existing rigid registration pipelines~\cite{Holz2015,Zhou2016}.


Deep networks have been applied to matching image patches and learning local image descriptors~\cite{Balntas2016,Han2015,Simo-Serra2015,Yi2016,ZagoruykoKomodakis2015,Zbontar2016}. Our research is informed by this work and applies related techniques to a different domain: point cloud registration. In particular, we investigate the effect of output dimensionality on accuracy and show that extremely low-dimensional descriptors can effectively represent the local geometry in an unstructured point cloud, significantly accelerating correspondence search in point cloud registration.

Learning has also been applied to shape classification and retrieval.
Researchers have considered volumetric~\cite{Wu2015,MaturanaScherer2015} and multi-view representations~\cite{Su2015}. These works do not deal with local geometric features and do not address the challenge of obtaining a local feature that is both accurate and compact. The difference between learning local geometric features and shape classification/retrieval is analogous to the difference between learning local image features~\cite{Simo-Serra2015} and image classification/retrieval~\cite{He2016,Babenko2014}.

\section{Overview}
\label{doc:overview}

\mypara{Parameterization.}
We parameterize the input to our model using spherical histograms centered at each point. These spherical histograms capture the local geometry in a neighborhood around each point. To incorporate rotational invariance, each spherical histogram is oriented to the normal and tangent spaces at each point. The interior of these spheres is subdivided along the radial, elevation, and azimuth directions. All neighboring points in the sphere are accumulated into the bins of the subdivision. The input parameterization is described in Section~\ref{sec:input}.

\mypara{Feature embedding.}
We train a deep network to map from the high-dimensional space of spherical histograms to a very low-dimensional Euclidean space. The network learns an embedding into a low-dimensional feature space that maps similar geometric neighborhoods to nearby points. The model is trained using the triplet embedding loss. This is described in Sections~\ref{sec:embedding} and~\ref{sec:training}.

\mypara{Correspondences.}
Given a mapping $f$ from a point into our learned feature space, computing correspondences between two point clouds $\P_i$ and $\P_j$ reduces to performing nearest-neighbor queries. We compute the set of features $f(\P_i)$ and $f(\P_j)$, and construct a $k$-d tree $\T$ on the point set $f(\P_j)$. For each point in $f(\P_i)$, we compute its nearest neighbor in $f(\P_j)$ using $\T$. As demonstrated in Section~\ref{doc:results}, correspondences computed using our feature space are much more accurate than correspondences computed using prior geometric feature descriptors. The low dimensionality of our features enables nearest-neighbor queries that are much faster than the second most accurate feature descriptor on real-world data.

\mypara{Applications.}
Our features can serve as drop-in replacements for existing descriptors. We demonstrate this by replacing widely used Fast Point Feature Histograms (FPFH)~\cite{Rusu2009} in existing geometric registration pipelines. This yields higher registration accuracy with no other modifications.
These experiments are reported in Section~\ref{doc:results}.


\section{Input Parameterization}
\label{sec:input}

Our basic approach is to start with a very high-dimensional representation of the raw local geometry around a point and train a deep network to embed this initial representation into a compact Euclidean space. Forming the initial representation is not trivial. Unlike images, which are laid out on a regular grid with a clear parameterization, a point cloud constitutes a set of unorganized points in $\Re^3$. Even the cardinality of the set of points within a given neighborhood is not fixed. One possibility is to discretize the input into a uniform voxel grid, but such representations are wasteful. For a $3$-dimensional grid with $C^3$ cells, a smooth $2$-dimensional surface will only intersect $O(C^2)$ cells: the rest are empty \cite{Itoh1995}. An alternative is to assume a clean parameterization of the underlying surface \cite{Boscaini2016}, but such a parameterization is not available in general.


Our initial representation is a histogram of the distribution of points in a local neighborhood, binned along a non-uniform radial grid \cite{Frome2004}.
Consider $\pp \in \P$ and let $\S$ be a sphere centered at $\pp$ with radius $r$. For rotational invariance, we estimate the normal $\nn_{\pp}$ and a local reference frame based on this normal~\cite{Salti2014}. Consider the third vector $\zz_{\pp}$ of the estimated local reference frame. If the dot product $\langle \nn_{\pp}, \zz_{\pp}\rangle < 0$, we flip the signs of all three vectors in the local reference frame.

\begin{figure}[!t]
\centering
\includegraphics[width=0.7\linewidth]{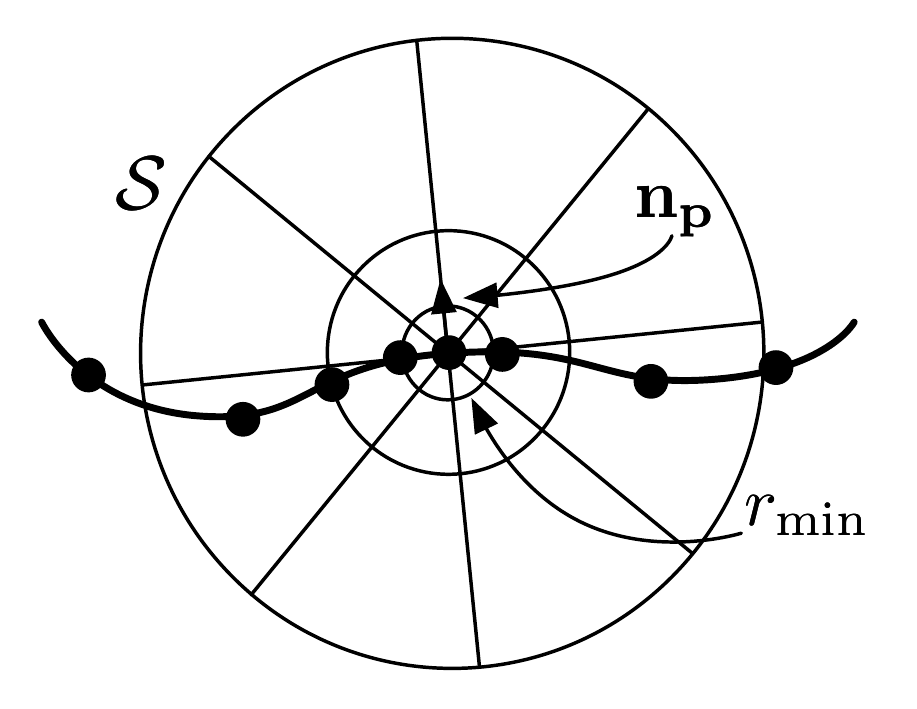}
\caption{Our input parameterization, illustrated in two dimensions for clarity. The sphere $\S$ is centered at the point $\pp$. In this two-dimensional illustration, the interior of $\S$ is subdivided into three bins along the radial direction and eight bins along the polar direction. This yields a 24-bin histogram into which the points in $\S$ are accumulated. The subdivision is aligned to the normal $\nn_{\pp}$. In the real three-dimensional setting, the histogram has approximately two thousand bins.}
\label{fig:input}
\end{figure}

The volume bounded by $\S$ can be subdivided into bins along the radial, elevation, and azimuth directions. These directions are defined in terms of the local reference frame. We subdivide the azimuth direction into $A$ bins, each of extent $2\pi/A$. The elevation direction is subdivided into $E$ bins, each of extent $\pi/E$. The radial direction, which has total span $r$, is logarithmically subdivided into $R$ bins with the following thresholds:
\begin{equation}
r_{i} = \exp\left(\ln{r_{\min}} + \frac{i}{R}\ln\left(\frac{r}{r_{\min}}\right)\right).
\end{equation}
The first threshold $r_0$ evaluates to $r_{\min}$, which avoids excessive binning near the center. The thresholds grow exponentially, yielding an initial representation of multi-scale context. The result is a spherical histogram with ${N = R\times E\times A}$ bins. This is illustrated in Figure \ref{fig:input}.

Let $\N \subset \P$ be the set of neighboring points that lie inside the sphere $\S$. The set $\N$ can be found efficiently using a $k$-d tree. For each point $\qq \in \N$, we locate the histogram bin that contains $\qq$ in constant time and increment the corresponding histogram value. After binning all the points in $\N$, we normalize the histogram by dividing each entry by $|\N|$. This yields a normalized $N$-dimensional feature vector that is used as input for a nonlinear embedding into a lower-dimensional Euclidean space.

\section{Feature Embedding}
\label{sec:embedding}

We train a deep network $f:\Re^N \rightarrow \Re^n$ to map the space of input histograms into a lower-dimensional Euclidean space $\Re^n$. This mapping serves two purposes. First, Euclidean distances between input histograms in $\Re^N$ are to a significant extent arbitrary and do not appropriately reflect the similarity or dissimilarity of the geometric contexts represented by the histograms. Second, nearest-neighbor search in the lower-dimensional space $\Re^n$ is much faster, which is important because nearest-neighbor search dominates the runtime of geometric registration pipelines~\cite{Elseberg2012,Zhou2016}.

The mapping is trained to pull similar features together while pushing dissimilar features apart. To this end, we use the triplet loss~\cite{SchultzJoachims2003,Weinberger2009}. This objective has been used to optimize feature embeddings for a number of applications in computer vision~\cite{Chechik2010,Schroff2015,Zbontar2016}.

Consider a set of triplets of input histograms ${\T = \{(\xx_{i}^{a}, \xx_{i}^{p}, \xx_{i}^{n})\}_{i}}$. Vector $\xx_{i}^{a}$ is referred to as the anchor of triplet $i$, vector $\xx_{i}^{p}$ is a positive example that is known to be similar to the anchor, and vector $\xx_{i}^{n}$ is a negative example that is known to be dissimilar. Given such a set of triplets, we optimize the following objective:
\begin{multline}
\cL(\btheta) = \frac{1}{|\T|} \sum_{i=1}^{|\T|} \big[ \|f(\xx_{i}^{a};\btheta) - f(\xx_{i}^{p};\btheta) \|^2 \\
 - \| f(\xx_{i}^{a};\btheta) - f(\xx_{i}^{n};\btheta)\|^2 + 1\big]_{+},
\label{equ:triploss}
\end{multline}
where $\btheta$ are the parameters of the mapping $f$ and $[ \cdot ]_{+}$ denotes $\max(\cdot, 0)$. Intuitively, $f$ is optimized such that $\xx_{i}^{a}$ is embedded closer to $\xx_{i}^{p}$ than to $\xx_{i}^{n}$, with a margin separating the distances.

We use a fully-connected network $f$ with 5 hidden layers. Each hidden layer contains 512 nodes and is followed by an elementwise truncation $\max(\cdot, 0)$. We validated our model architecture with a controlled experiment reported in the supplement. At test time, computing the $n$-dimensional descriptor corresponding to an input histogram amounts to a sequence of matrix multiplications and elementwise operations.




\section{Training}
\label{sec:training}

Consider a set of point clouds $\{\P_{i}\}_i$ that depict overlapping fragments of a scene. Let $\{\TT_i\}_i$ be a set of rigid transformations that align the point clouds $\{\P_{i}\}_i$. Thus $\{\TT_i \P_{i}\}_i$ is a set of point clouds aligned to a common coordinate frame, in which distances between points $\pp \in \TT_i \P_i$ and $\qq \in \TT_j \P_j$ that depict nearby points in the latent scene are small. In this section we assume that the point clouds $\{\P_{i}\}_i$ and transformations $\{\TT_{i}\}_i$ are given. Data in this form can be obtained from a variety of sources including scene reconstruction pipelines.





Consider a single point cloud $\P_{i}$ and a point \mbox{$\pp \in \P_{i}$}. Let $\operatorname{nn}(\pp, \P_{i})$ denote the nearest neighbor of $\pp$ in \mbox{$\P_{i} \setminus \pp$}. Let $\eps_i$ be the median of the set of distances \mbox{$\{\|\pp - \operatorname{nn}(\pp, \P_i)\| : \pp \in \P_{i}\}$} and define $\eps = \max_i \eps_i$.

Now consider a pair of point clouds $(\P_{i}, \P_{j})$. For each point $\pp \in \TT_{i}\P_{i}$ we can compute the nearest neighbor $\operatorname{nn}(\pp, \TT_{j}\P_{j})$ of $\pp$ in $\TT_{j}\P_{j}$. Consider the fraction of such pairs that are within distance $\eps$. Specifically, define
\begin{equation}
\alpha_{i,j} = \frac{\left|\{\pp \in \TT_{i}\P_{i}: \|\pp - \operatorname{nn}(\pp, \TT_{j}\P_{j})\| \leq \eps\}\right|}{|\P_{i}|}
\end{equation}
and similarly for $\alpha_{j, i}$. We say that $\P_{i}$ and $\P_{j}$ overlap if \mbox{$\min(\alpha_{i,j},\alpha_{j,i}) \geq 0.3$}. This implies that the underlying surfaces from which $\P_{i}$ and $P_{j}$ were sampled overlap by at least $30\%$.

\begin{figure}[!t]
\centering
\includegraphics[width=0.8\linewidth]{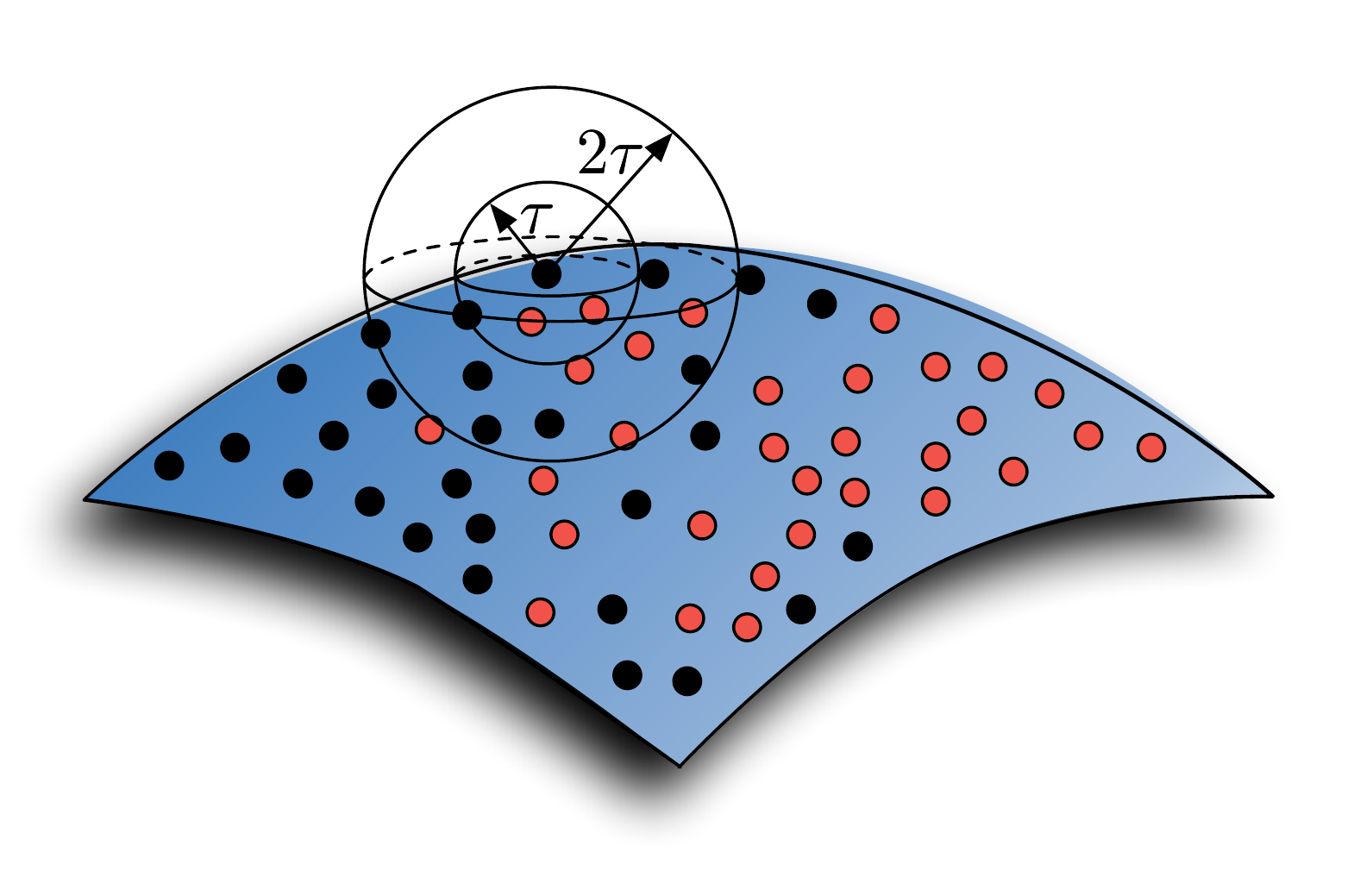}
\caption{Two overlapping points clouds, shown here in red and black, are sampled from an underlying surface. We consider two concentric spheres of radius $\tau$ and $2\tau$ around a black point $\pp$. The red points in the innermost sphere, which form the set $\N_{\pp, \tau}$, are good correspondences for $\pp$. The red points in the outermost sphere form the set $\N_{\pp, 2\tau}$. We generate triplets for $\pp$ by sampling $\xx^{p} \in \N_{\pp, \tau}$ and $\xx^{n} \in \N_{\pp, 2\tau} \setminus \N_{\pp, \tau}$.}
\label{fig:triplets}
\end{figure}

Consider the set $\OC$ of overlapping pairs of point clouds from $\{\P_{i}\}_i$. For each pair $(\P_i, \P_j) \in \OC$ we examine each point \mbox{$\pp \in \P_i$}. We compute the set of neighbors $\N_{\pp,\tau}^{j}$ in $\P_j$ that are at distance at most $\tau$ from $\pp$. When $\tau$ is sufficiently small, the points in $\N_{\pp, \tau}^{j}$ are good correspondences for $\pp$ in $\P_{j}$.
Similarly consider $\N_{\pp, 2\tau}^{j}$, the set of points in $\P_j$ that are at distance at most $2\tau$ from $\pp$. The set $\N_{\pp,2\tau}^{j} \setminus \N_{\pp,\tau}^{j}$ contains difficult negative examples for $\pp$. These points have local geometries that are in general more similar to that of $\pp$ than a randomly chosen point, but are not as close as those in $\N_{\pp,\tau}^{j}$. This is illustrated in Figure~\ref{fig:triplets}.

We generate training triplets \mbox{$(\xx^{a}, \xx^{p}, \xx^{n})$} by sampling a pair of point clouds $(\P_i, \P_j) \in \OC$, sampling a point \mbox{$\xx^a \in \P_i$} from the overlap region of $\P_i$ and $\P_j$, sampling $\xx^p$ from the set $\N_{\xx^a, \tau}^{j}$, and sampling $\xx^n$ from the set $\N_{\xx^a,2\tau}^{j} \setminus \N_{\xx^a,\tau}^{j}$.
This procedure is used to generate a large number of training triplets. The triplets are then permuted randomly and partitioned into minibatches.

We use minibatches of size $512$ and train the mapping $f$ using Adam~\cite{KingmaBa2015}. The initial weights of the hidden nodes are drawn from a normal distribution with mean $0$ and standard deviation $0.1$. The learning rate is set to $10^{-4}$. The parameters for the exponential decay of the first and second moment estimates are set to $\beta_1 = 0.9$ and $\beta_2 = 0.999$. The network is trained for three epochs.

\section{Experiments}
\label{doc:results}

\subsection{Setup}

\mypara{Input parameterization.}
For the input parameterization, we use $R=17$ subdivisions in the radial direction, $E=11$ in the elevation direction, and $A=12$ in the azimuth direction. The dimensionality of the input histogram is thus $N=2@244$. We validate these choices via controlled experiments that are reported in the supplement.

Our approach yields a family of features parameterized by dimension. The primary setting of our feature space dimensionality is $n=32$; the corresponding feature is referred to as \mbox{CGF-32}. The experimental results will demonstrate that this low dimensionality significantly outperforms prior, much larger descriptors.

On laser scan data, the radius $r$ of the sphere $\S$ is set to $17\%$ of the diameter of each model and $r_{\min}$ is set to $1.5\%$ of the diameter. The value of the search radius is validated via controlled experiments reported in the supplement. The local reference frame at each point is computed using a search radius of $2\%$ of the diameter.

On data from SceneNN~\cite{Scenenn2016}, which has absolute metric scale, the radius $r$ is set to 1.2 meters, which is approximately $17\%$ of the diameter of the grid of each fragment, and $r_{\min}$ is set to 0.1 meters. The local reference frame is computed using a search radius of 0.25 meters.

\mypara{Laser scan data.}
For experiments with laser scan data, we use a number of public-domain 3D models that are commonly used for this purpose. We use three models from the AIM@SHAPE repository (Bimba, Dancing Children, and Chinese Dragon), four models from the Stanford 3D Scanning Repository (Armadillo, Buddha, Bunny, and Stanford Dragon), and the Berkeley Angel~\cite{Kolluri2004}. Four of these models~-- Angel, Bimba, Bunny, and Chinese Dragon~-- were used as the training set, the Dancing Children model was used for validation, and the remaining three models~-- Armadillo, Buddha, and Stanford Dragon~-- were used as the test set.

For each model in the training and validation sets~-- Angel, Bimba, Bunny, Chinese Dragon, and Dancing Children~-- we synthesize depth images from 14 views uniformly distributed along the surface of an enclosing sphere. For each depth image we construct a point cloud that lies on the model. We compute the set of pairs of point clouds $\OC$ that overlap in world space by at least $30\%$. Since some of these models do not have absolute scale, we set parameters and measure precision in relation to the diameter of the model. Synthesizing depth images allows us to automatically generate as much training data as we need and provides a controlled training environment in which we can validate our design choices. We found that descriptors trained on such synthetically scanned models successfully generalize to raw laser scans.

For testing we use the original raw laser scans of the models in our test set -- Armadillo, Buddha, and Stanford Dragon. All three models were scanned with a Cyberware 3030 MS scanner. Armadillo has 114 scans, Buddha has 58 scans, and the Stanford Dragon has 71 scans. Using the provided alignments we compute a set of pairs of scans $\OC$ that overlap in world space by at least $30\%$.  These models demonstrate the ability of CGF to generalize to new domains, handle symmetric objects, and cope with noise encountered in laser scanned models.

\mypara{SceneNN data.}
For experiments on real indoor scenes, we use SceneNN~\cite{Scenenn2016}, a comprehensive recent dataset of indoor scenes scanned with consumer depth cameras. Starting from the raw SceneNN scans, we create fragments and register them using the pipeline of Choi et al.~\cite{Choi2015}. Each fragment is fused from 100 consecutive frames.

50\% of the scenes are used for training, 25\% for validation, and 25\% as the test set, split randomly. We will publish our train/val/test split so that others can replicate our experiments. Let $\OC$ be the set of pairs of overlapping fragments in the training scenes. For maximally precise alignment during training, we refined the registration of each pair $(\P_i, \P_j) \in \OC$ using ICP \cite{Besl1992}. We use the implementation of ICP provided in the Point Cloud Library \cite{Holz2015}.


\mypara{Training.}
For each point cloud in the training set (synthetic depth image in the case of laser scan data, scene fragment in the case of SceneNN), we sample 40 triplets per point. Of these 40 triplets, 15 are constructed by sampling negatives from ${\N_{\pp, 2\tau} \setminus \N_{\pp, \tau}}$, as described in Section~\ref{sec:training}. The remaining 25 are constructed by sampling negatives from the entire model. The threshold $\tau$ is set to $1\%$ of the model's diameter in the case of laser scans and 7.5 cm in the case of SceneNN.

\mypara{Baselines.}
We compare CGF to six well-known local descriptors: Point Feature Histograms (PFH)~\cite{Rusu2008-IROS} (dimensionality 125),  Fast Point Feature Histograms (FPFH)~\cite{Rusu2009} (dimensionality 33), Rotational Projection Statistics (RoPS)~\cite{Guo2013} (dimensionality 135), Signature of Histogram Orientations (SHOT)~\cite{Salti2014} (dimensionality 352), Spin Images (SI)~\cite{JohnsonHebert1999} (dimensionality 153), and Unique Shape Contexts (USC)~\cite{Tombari2010b} (dimensionality 1,980). For RoPS we use the implementation provided by the authors~\cite{Guo2013}. For all other baselines we use the implementations provided in the Point Cloud Library~\cite{Holz2015}.
Each of these existing geometric feature descriptors has several parameters that need to be tuned to ensure good performance. We performed extensive hyperparameter sweeps to ensure that each baseline performed as well as possible in our experiments.

We have also applied Principal Components Analysis (PCA) to embed our input 2,244-dimensional histograms into $\Re^n$, using our primary dimensionality $n=32$. This evaluates the advantage of the presented nonlinear feature embedding over a linear embedding of the same input into the same space.

Additional baselines and controlled experiments are reported in the supplement.

\begin{figure*}[!t]
\centering
\begin{minipage}{0.97\textwidth}
\centering
\begin{tabular}{c c c}
    \includegraphics[width = .43\linewidth]{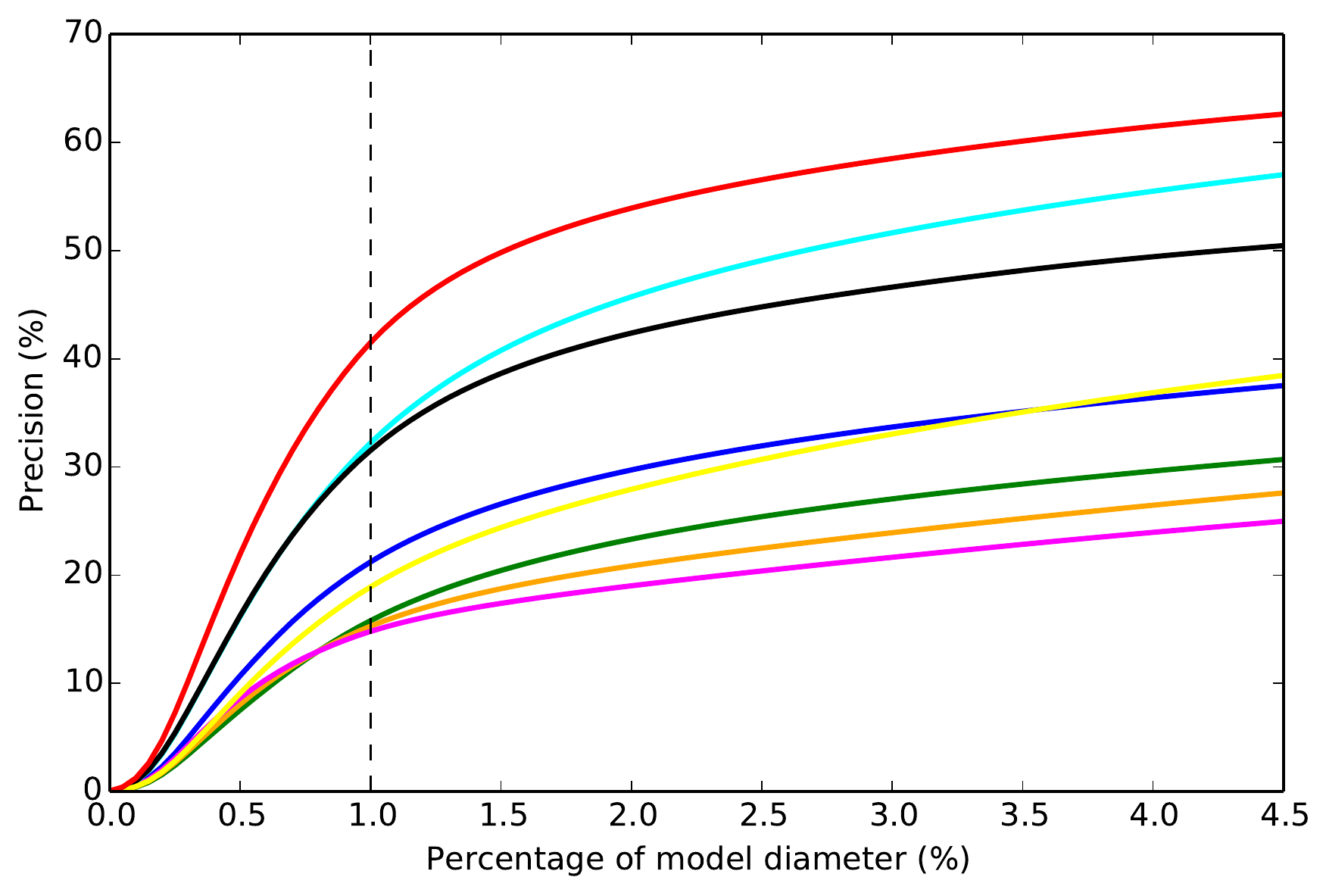} &
    \includegraphics[width = .43\linewidth]{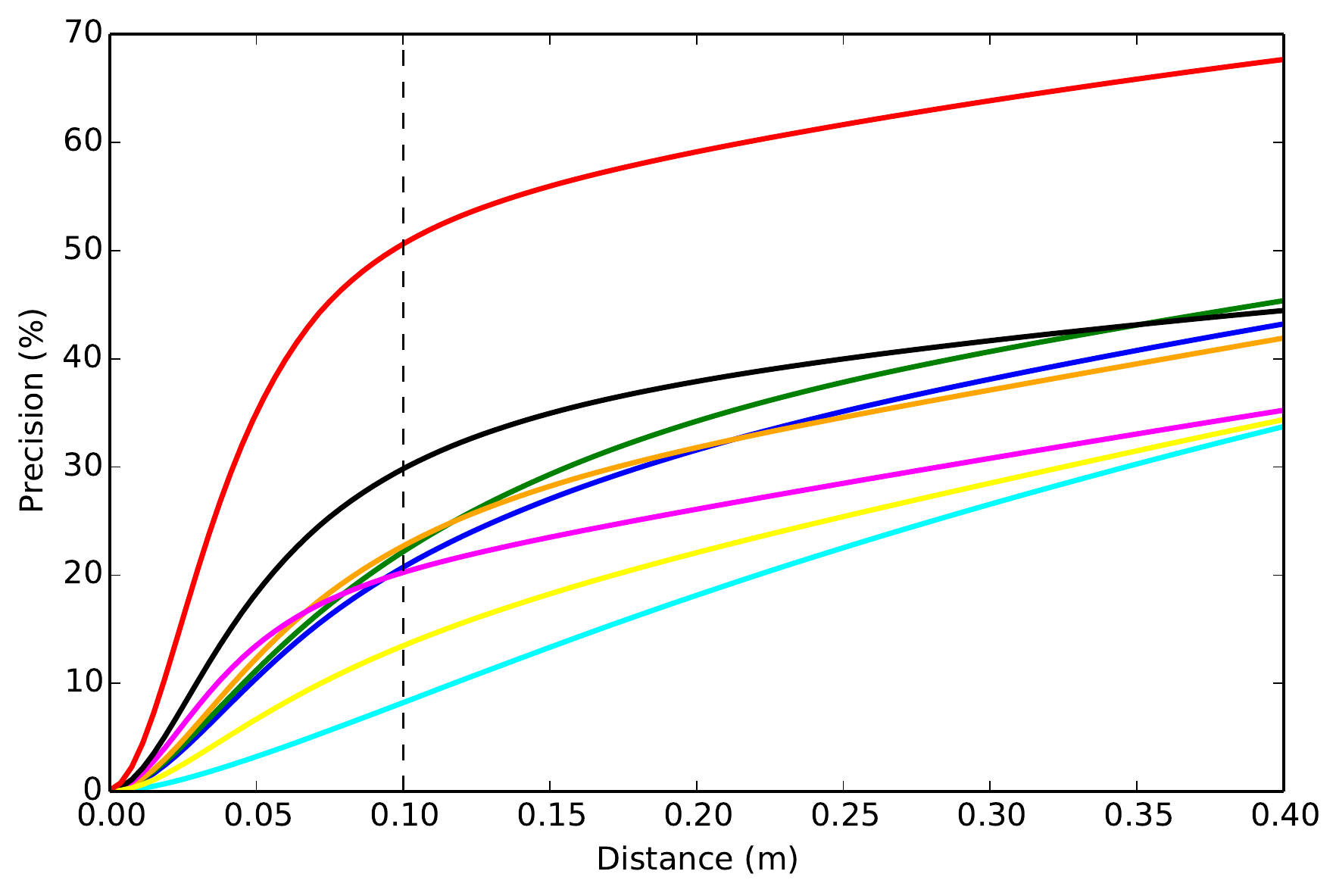} &
    \includegraphics[trim= 0 -1cm 0 0, width = .11\linewidth]{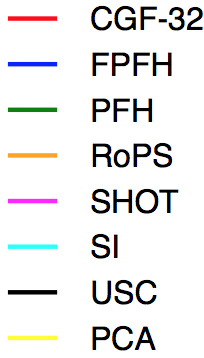}\\
     \small{(a) Precision on laser scan data test set} & \small{(b) Precision on the SceneNN test set} & \\
\end{tabular}
\end{minipage}\hfill
\vspace{1mm}
\caption{(a) The precision of several geometric feature descriptors on laser scan data in the test set. For correspondences provided by CGF-32, 41.4\% are precise to within $1\%$ of the diameter. Prior feature descriptors are less accurate. (b) Precision of local geometric features on pairs of fragments from the SceneNN test set. CGF-32 yields the highest precision: 50.6\% of the matches computed in the learned feature space are within 10 cm of the ground truth. USC (a \mbox{1,980-dimensional} descriptor) comes in second at 29.8\%.}
\label{fig:precision}
\vspace{-1mm}
\end{figure*}

\begin{figure*}[!t]
\centering
\begin{minipage}{0.97\textwidth}
\centering
\begin{tabular}{c c c}
     \includegraphics[width = .43\linewidth]{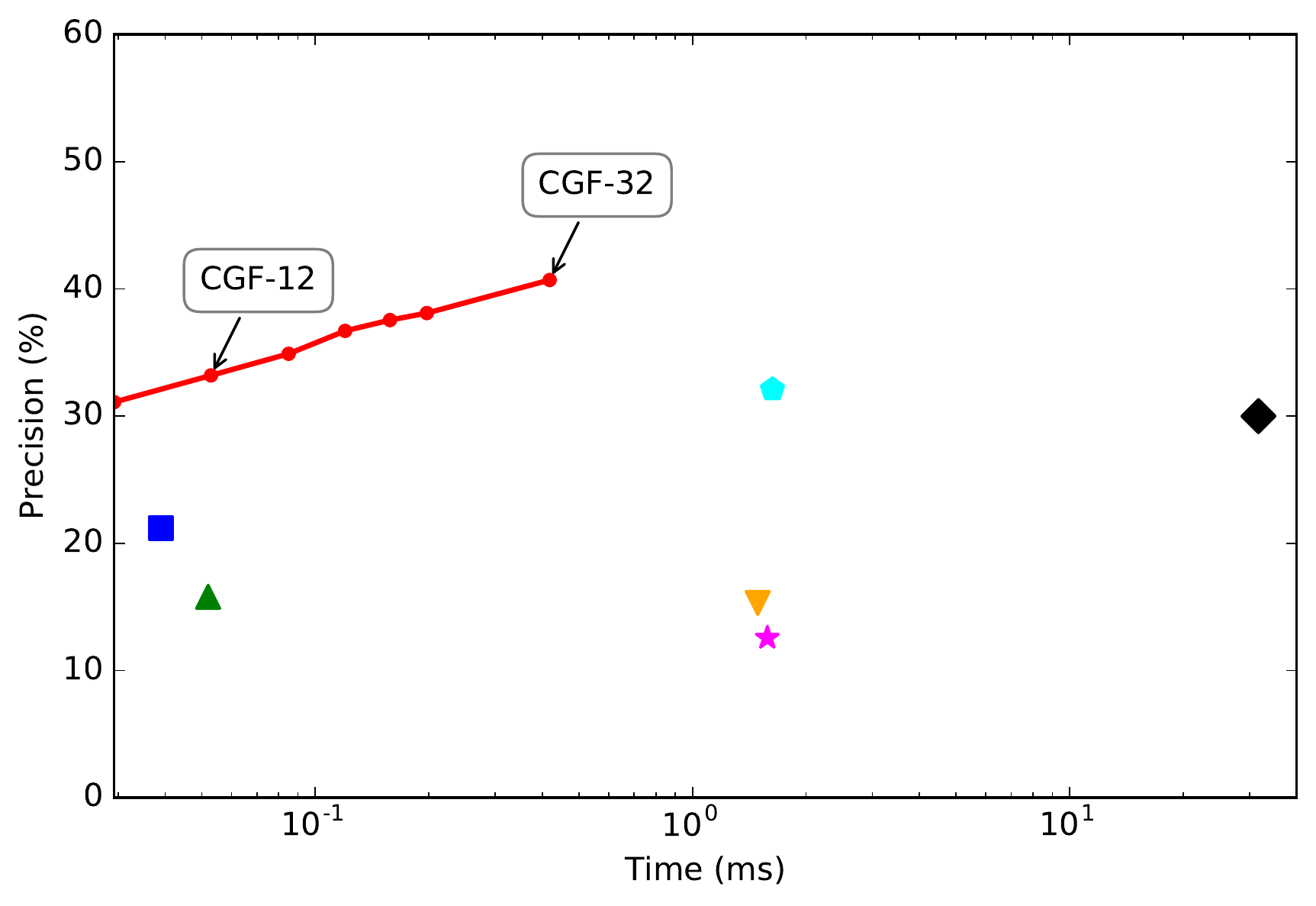} &
     \includegraphics[width = .43\linewidth]{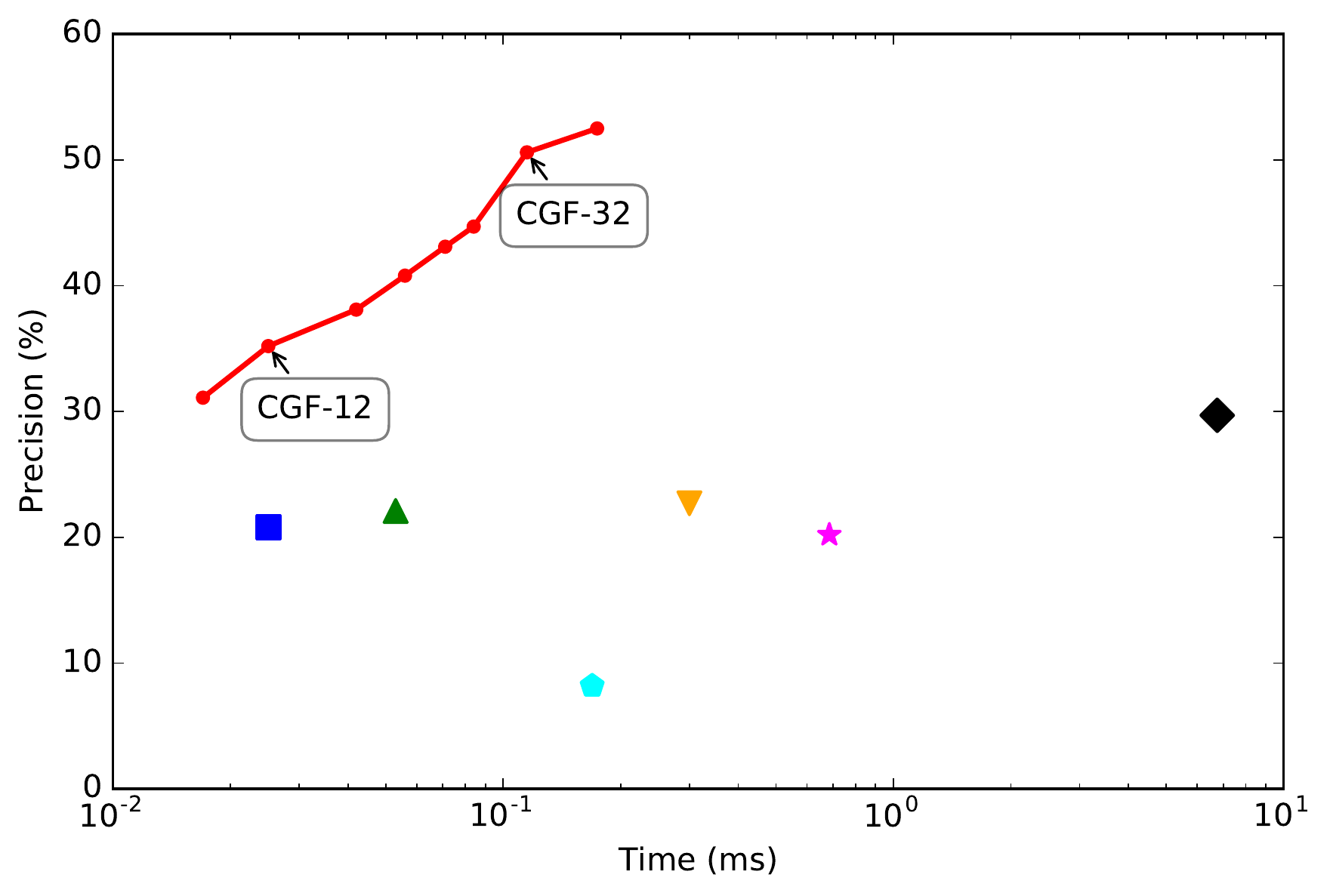} &
     \includegraphics[trim=-0 -1.3cm 0 0, width = .1\linewidth]{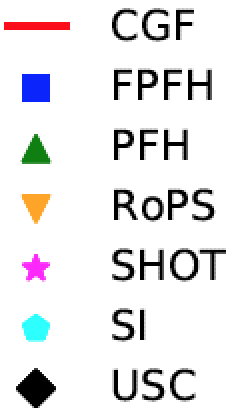}\\
    \small{(a) Query time and precision on laser scan data test set} & \small{(b) Query time and precision on the SceneNN test set} & \\
\end{tabular}
\end{minipage}\hfill
\vspace{1mm}
\caption{(a) The query time and precision of several geometric feature descriptors on laser scan data in the test set. CGF-32 has an average query time of 0.42 ms, 3.9 times faster than the second most accurate feature (SI, 1.62 ms). (b) The query time and precision of local geometric features on pairs of fragments from the SceneNN test set. CGF-32 has an average query time of 0.1 ms, 67 times faster than the second most accurate feature (USC, 6.75 ms). The horizontal axis (time) is on a logarithmic scale.}
\label{fig:timings}
\vspace{-1mm}
\end{figure*}

\mypara{Accuracy measure.}
Let $\{\P_{i}\}_{i}$, $\{\TT_{i}\}_{i}$, and $\OC$ be defined as in Section \ref{sec:training} and consider an overlapping pair $(\P_i, \P_j) \in \OC$. Given a function $f$ that maps points to geometric features, a set of correspondences between $\P_i$ and $\P_j$ can be found by first computing the sets of geometric features $f(\P_i)$ and $f(\P_{j})$. Then we build a $k$-d tree $\T$ on the set $f(\P_j)$. For each point $\pp \in \P_i$, we compute $\operatorname{nn}(f(\pp), f(\P_j))$ by performing a nearest neighbor query in $\T$. Define
\begin{equation}
\C_f = \{(\pp, \qq) : \pp \in \P_i, \qq \in \P_j, f(\qq) = \operatorname{nn}(f(\pp), f(\P_j))\}
\end{equation}
as the set of matches yielded by the feature $f$.

Since $\P_i$ only partially overlaps with $\P_j$, we first discard all correspondences $(\pp, \qq)$ such that
\begin{equation}
  \| \TT_i\pp - \operatorname{nn}(\TT_i \pp, \TT_j\P_j)\| > \tau.
\end{equation}
These points have no ground-truth correspondence in $\P_j$. Let $\C'_f$ denote the remaining set of correspondences.

For any distance threshold $x$, we can compute the fraction of matches that are within distance $x$ of the ground truth:
\begin{equation}
\pre_f(x) = \frac{|\{\| \TT_i \pp - \TT_j \qq\| \leq x : (\pp, \qq) \in \C'_f\}|}{|\C'_f|}.
\end{equation}
This will be our primary measure for evaluating the accuracy of different features $f$.

\mypara{Timings.}
Average correspondence search times for different descriptors were benchmarked using a single thread on an Intel Xeon E7-8890 2.5 GHz CPU. We use FLANN~\cite{flann2014} to perform nearest-neighbor queries.


\subsection{Laser scan data}
\label{sec:synthetic}

Precision of different features on the test set is shown in Figure~\ref{fig:precision}(a). CGF-32 is much more accurate than existing descriptors. For example, $41.4\%$ of the correspondences produced by CGF-32 lie within $1\%$ of the model's diameter of the true match, whereas the most precise prior feature, SI, yields only $32.2\%$ precision at this distance. CGF-32 improves over SI by $28.5\%$ in relative precision while being $4.7$ times more compact.

\begin{figure*}[!t]
\centering
\begin{minipage}{0.97\textwidth}
\centering
\begin{tabular}{c c c c c c c}
    \raisebox{8mm}{\rot{Laser scans}} &
    \includegraphics[width = .165\linewidth]{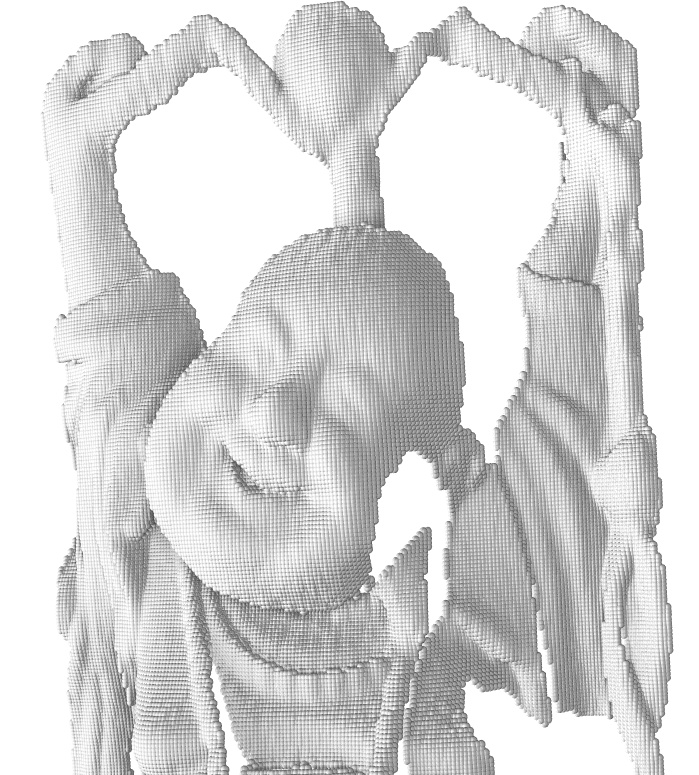} &
    \includegraphics[width = .165\linewidth]{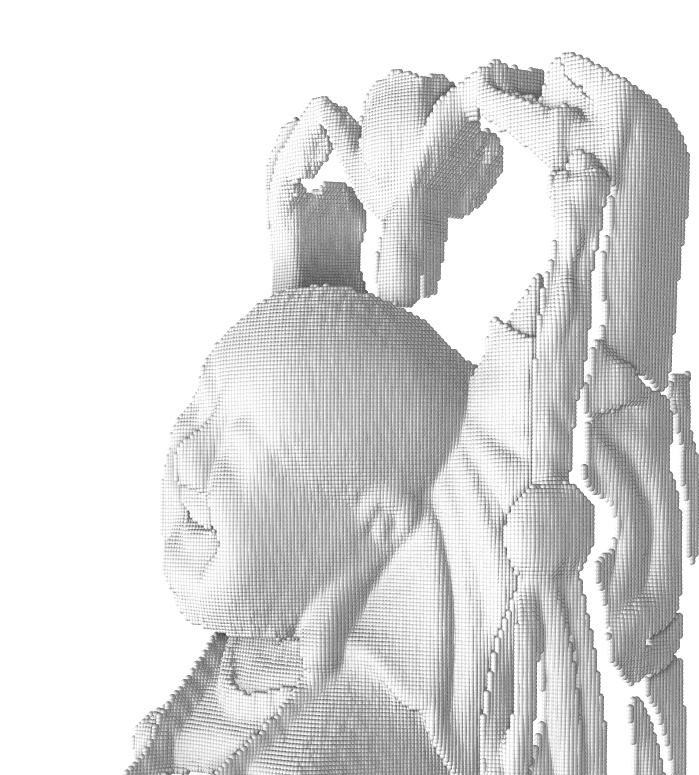} &
    \includegraphics[width = .165\linewidth]{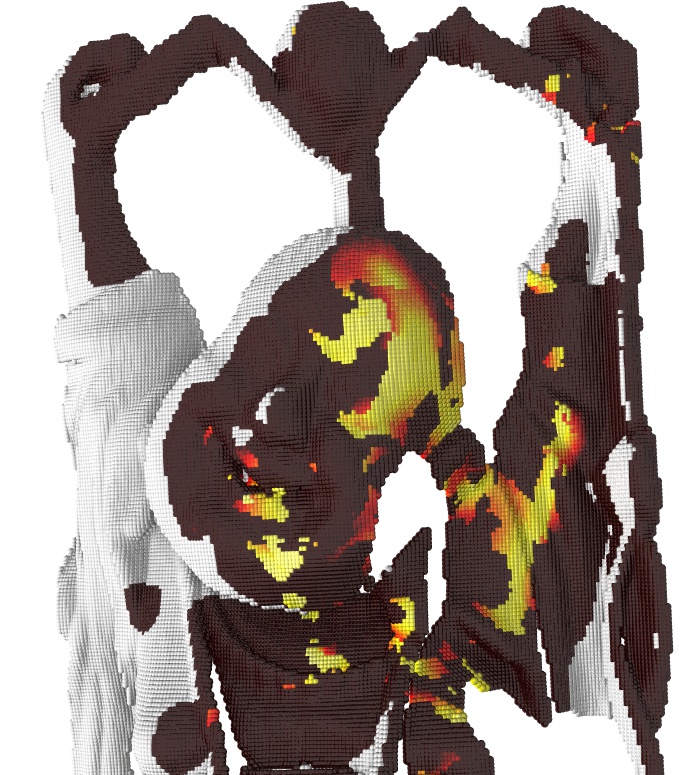} &
    \includegraphics[width = .165\linewidth]{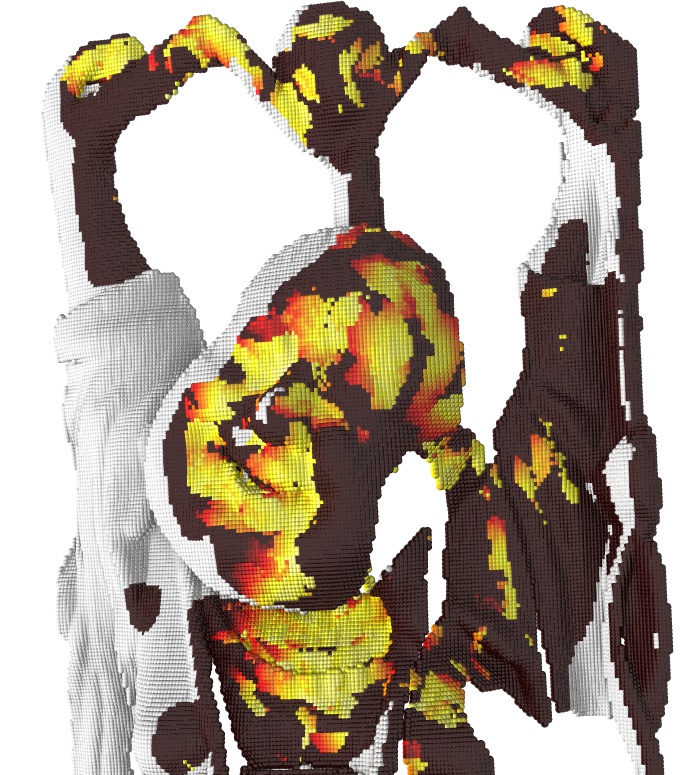} &
    \includegraphics[width = .165\linewidth]{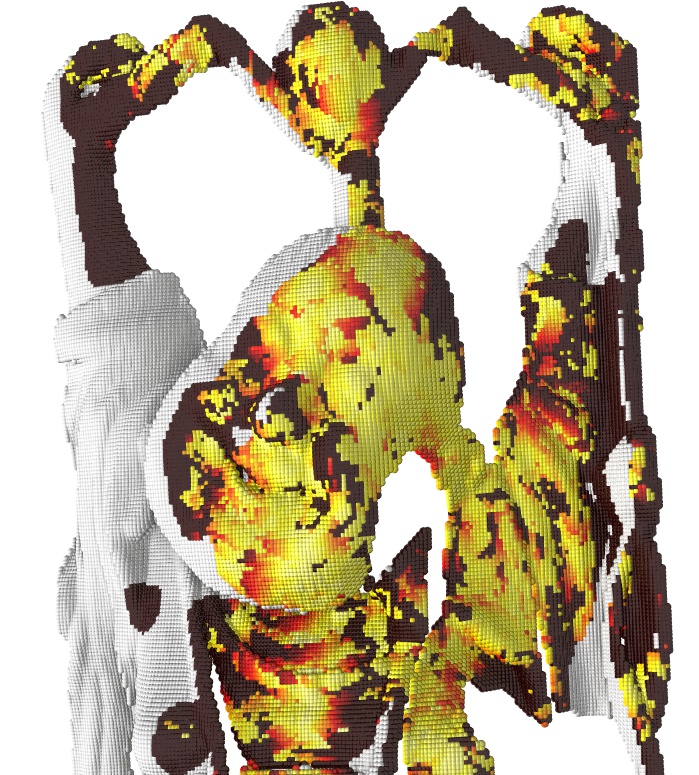} &
    \includegraphics[height= 0.2\linewidth]{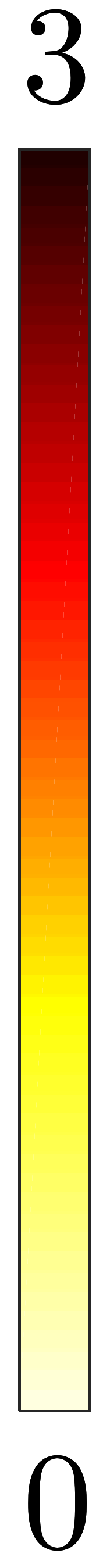}\\
    \raisebox{8mm}{\rot{SceneNN}} &
    \includegraphics[width = .165\linewidth]{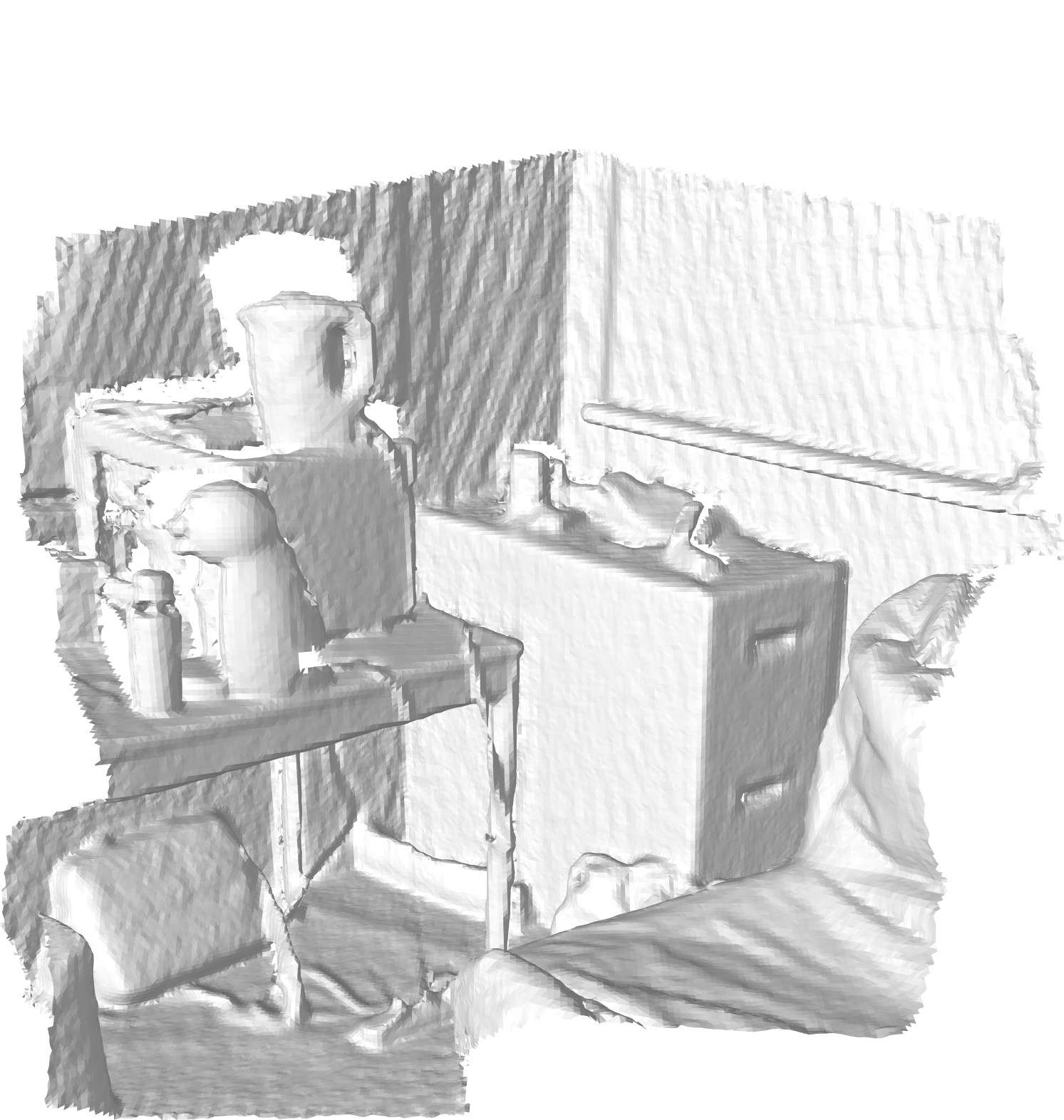} &
    \includegraphics[width = .165\linewidth]{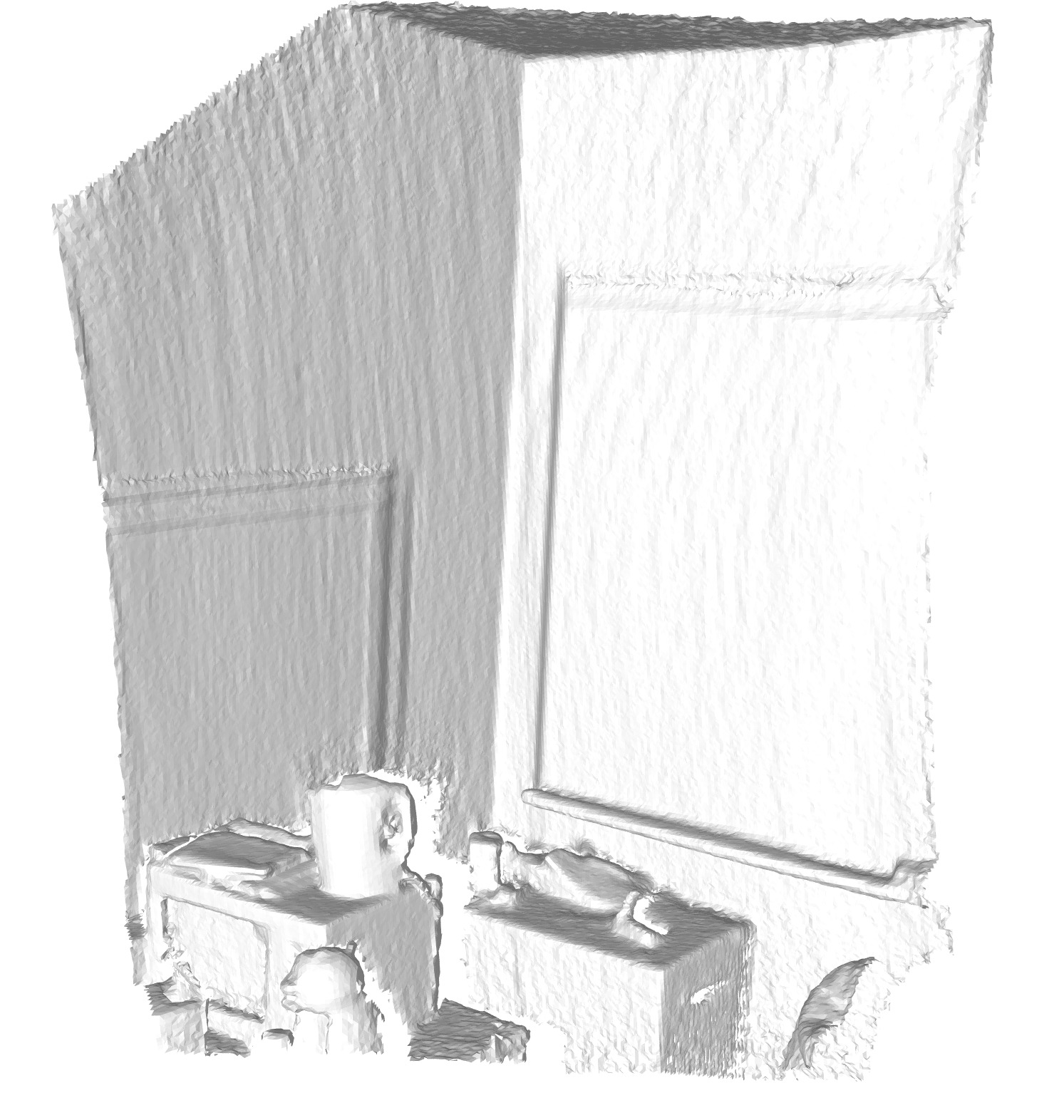} &
    \includegraphics[width = .165\linewidth]{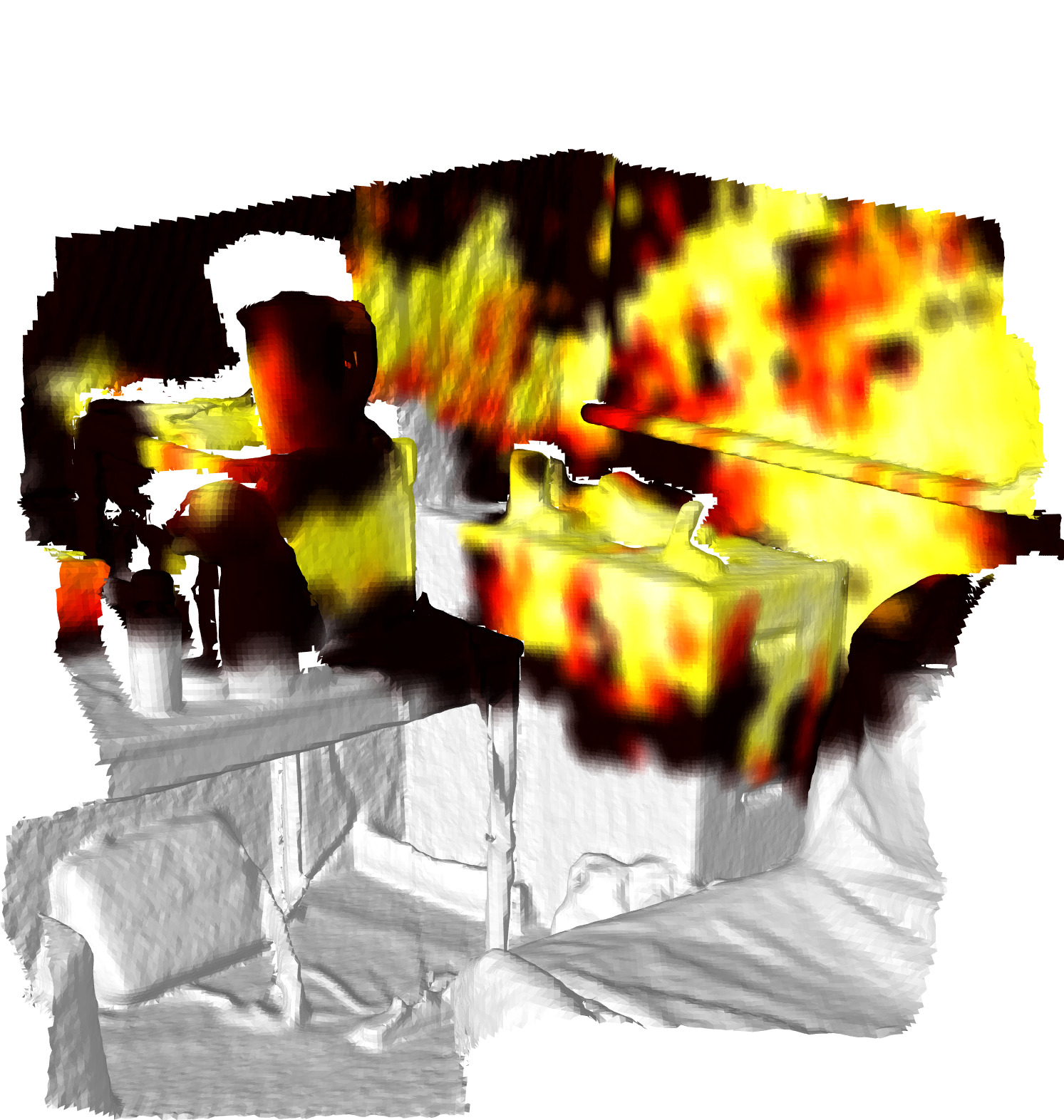} &
    \includegraphics[width = .165\linewidth]{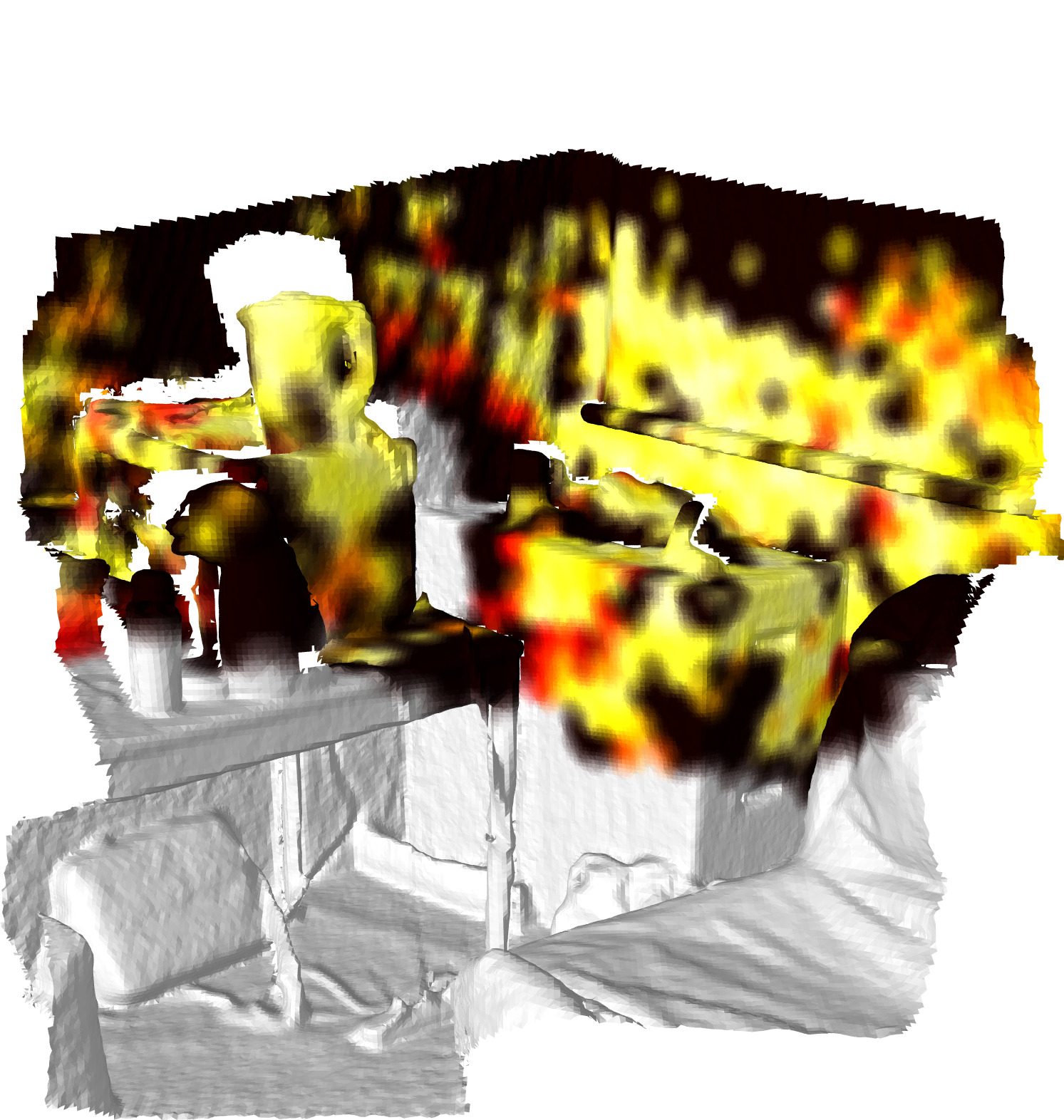} &
    \includegraphics[width = .165\linewidth]{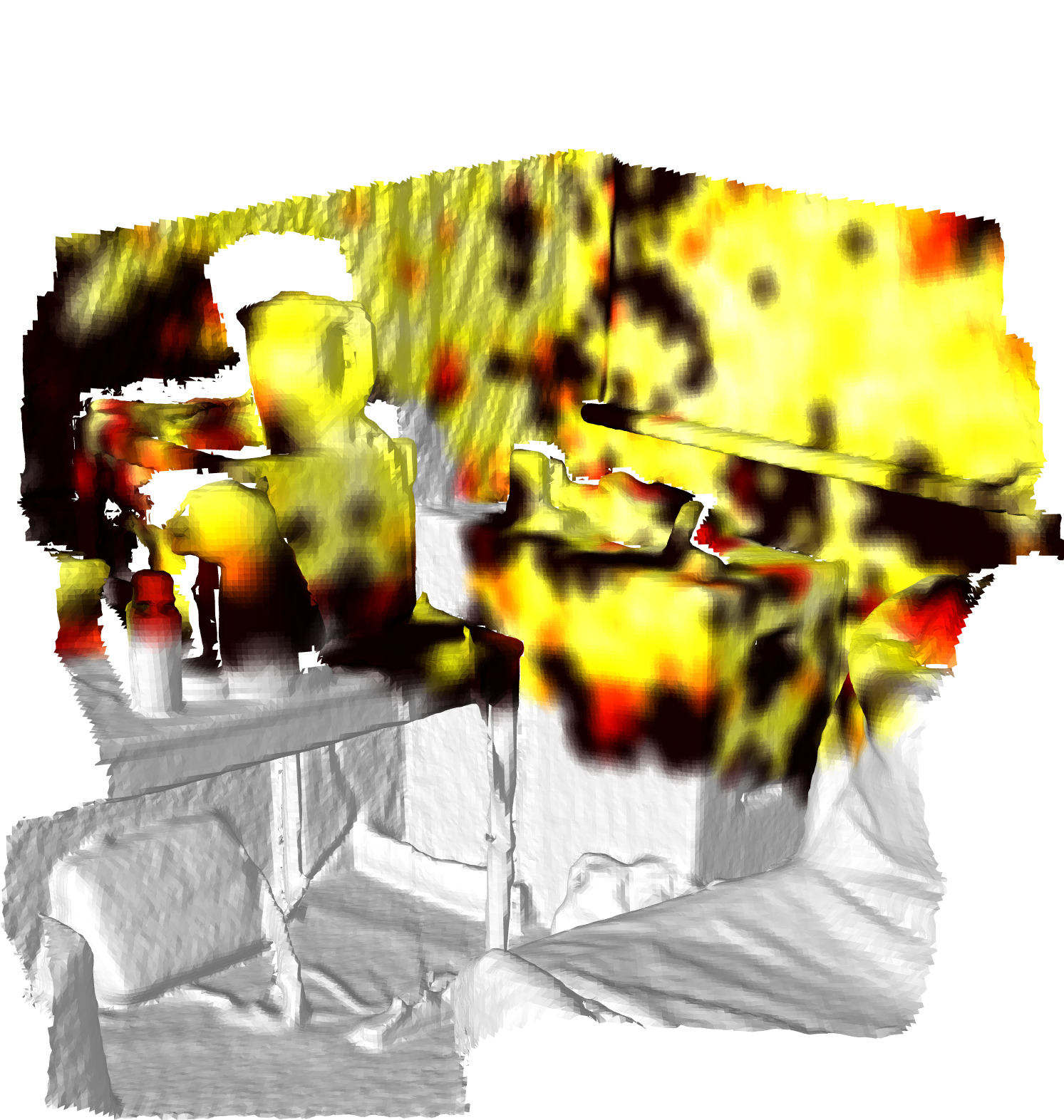} &
    \includegraphics[height = 0.2\linewidth]{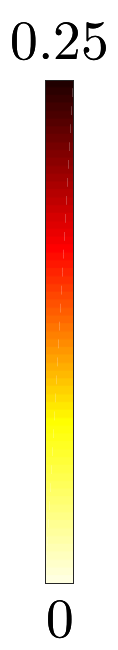}\\
    & \small{(a) source} & \small{(b) target} & \small{(c) FPFH} & \small{(d) USC} & \small{(e) CGF-32}\\
\end{tabular}
\end{minipage}\hfill
\vspace{1mm}
\caption{{\bf Top.} (a,b) Two laser scans of the Buddha statue. (c-e) Error magnitudes of matches established across the two scans in different feature spaces. CGF provides broad coverage of the surface with accurate matches. Units are in percentage of the model's diameter: black corresponds to error of 3\% of the diameter or higher.
{\bf Bottom.} (a,b) Two fragments in the SceneNN test set. (c-e) Error magnitudes of correspondences established across these fragments in different feature spaces. Black corresponds to errors of 25 cm or higher. Correspondences established via CGF are more precise on average. Note the thin structure above the large hole in the middle of the fragment, along which all other feature spaces fail to establish good correspondences.
Points shown in grey do not have a ground-truth correspondence on the other point cloud.
}
\label{fig:vis-error}
\vspace{-1mm}
\end{figure*}

\mypara{Timings.}
Query times for different features are presented in Figure~\ref{fig:timings}. CGF-32 has an average query time of $0.42$ ms, which is $3.9$ times faster than the second most accurate feature (SI, 1.62 ms) and $75$ times faster than USC (31.6 ms). CGF-12 has an average query time of $0.05$ ms, slightly slower than the fastest feature (FPFH, 0.04 ms) while being more precise than all baselines (33.2\% at $1\%$ of the diameter).

\mypara{Visualization.}
Figure~\ref{fig:vis-error} (top) shows error distributions of correspondences established in different feature spaces over two laser scans of the Buddha statue.

\subsection{SceneNN data}
\label{sec:scenenncorr}

Precision of different feature descriptors on real-world scene fragments from the SceneNN test set is shown in Figure~\ref{fig:precision}(b). 50.6\% of the matches established with CGF-32 are within 10 cm of the true match, much more than USC (29.8\%), RoPS ($22.7\%$), PFH ($22.1\%$), FPFH ($20.7\%$), SHOT ($20.2\%$), and SI ($8.2\%$).
The baseline constructed by applying PCA to our 2,244-dimensional input parameterization yielded precision of $13.4\%$, far lower than the precision of the learned nonlinear embedding into the same space. Note that the second highest performing feature on laser scan data, SI, performed poorest on SceneNN.

\mypara{Timings.}
Query times for different features are presented in Figure~\ref{fig:timings}. CGF-32 has an average query time of $0.1$ ms, $67$ times faster than the second most accurate feature (USC, $6.75$ ms). CGF-12 has an average query time of $0.025$ ms, matching the speed of FPFH. In addition to its speed, \mbox{CGF-12} is more precise than all other features, with $31.5\%$ of correspondences within 10 cm of the true match.

\mypara{Visualization.}
Figure~\ref{fig:vis-error} (bottom) shows error distributions of correspondences established in different feature spaces over two fragments in the SceneNN test set.

\subsection{Visualization}
\label{sec:visualization}
\begin{figure}[h]
\centering
\begin{tabular}{c c}
    \includegraphics[width = .45\linewidth]{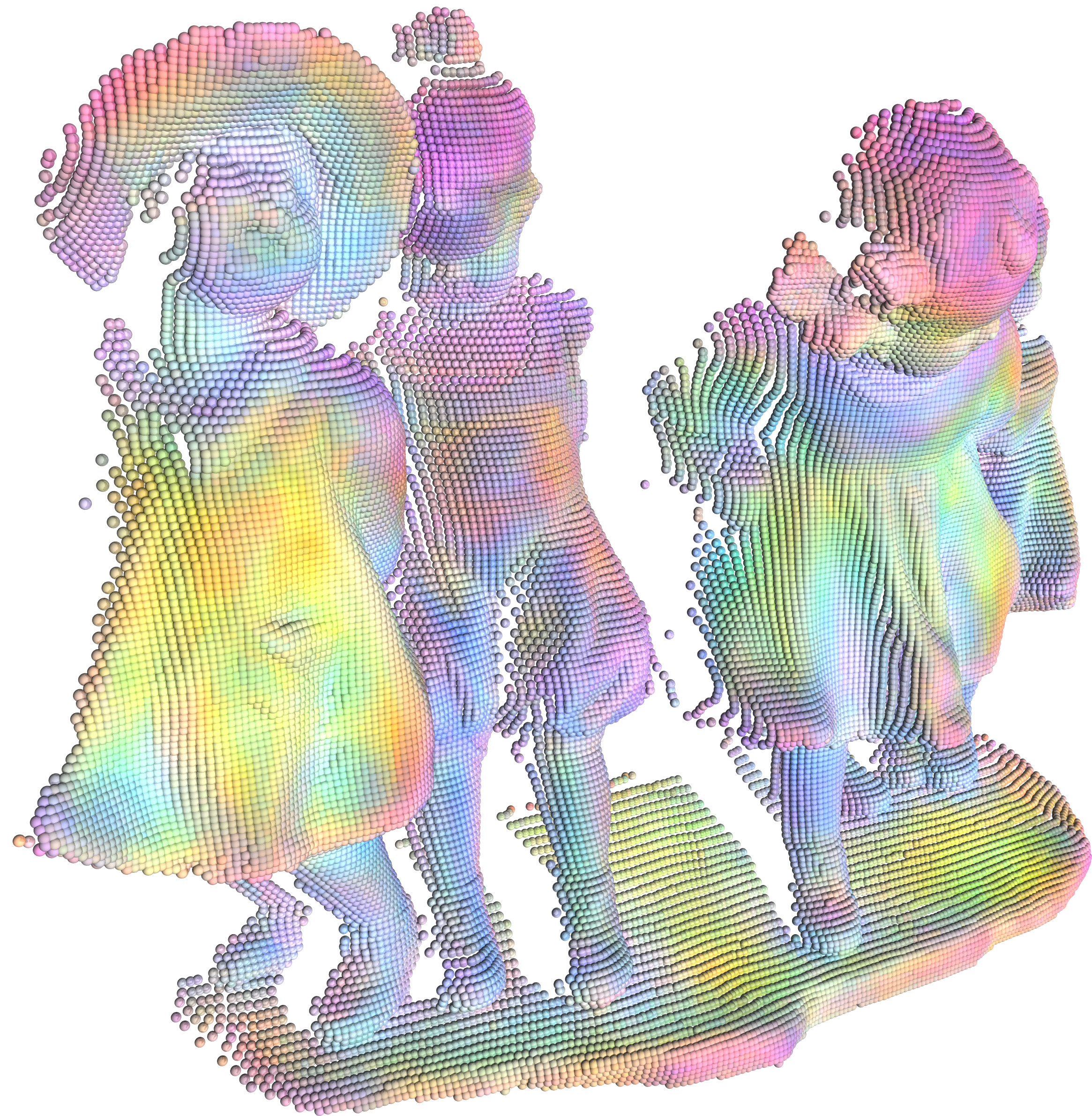} &
    \includegraphics[width = .45\linewidth]{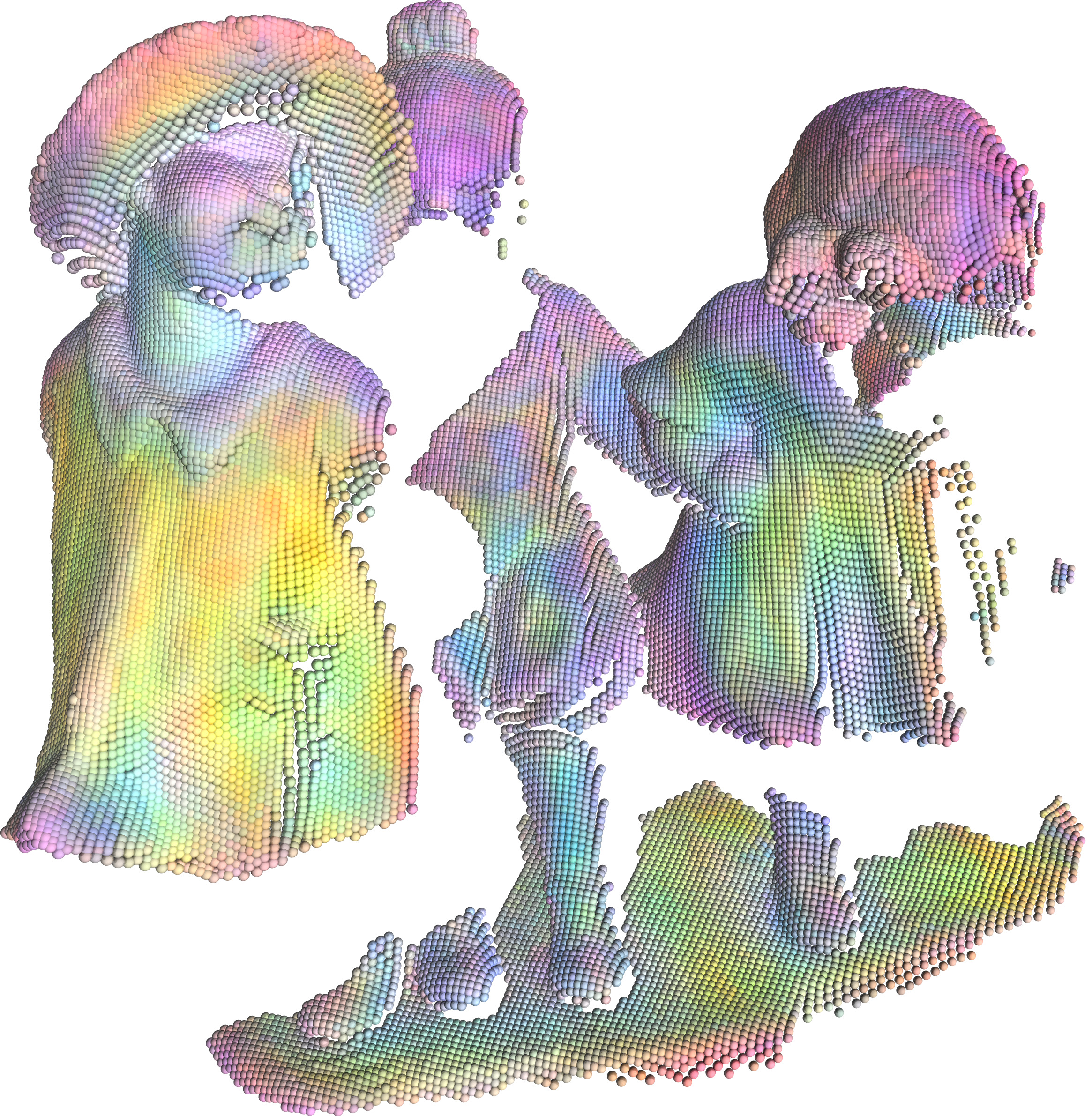}
\end{tabular}
\vspace{0.5mm}
\caption{Visualization of local features over two views of the Dancing Children model. Features were projected from the learned feature space into the RGB color space.}
\label{fig:dc-vis}
\end{figure}

To get a qualitative sense of the learned representation, we can visualize the variation of the learned features over the surface of any model. Specifically, we can use PCA to project from the learned feature space into the \mbox{3-dimensional} RGB color space. Given a point set, we can evaluate the learned feature for every point, use the learned linear mapping to obtain the corresponding color, and assign this color to the point. Figure \ref{fig:dc-vis} shows the result of this procedure for two synthesized views of the Dancing Children model. Note that the feature mapping appears stable, coherent, and discriminative. Corresponding points on the two views of the model tend to have similar color. Color varies more rapidly in regions of high-frequency geometric variation and is more stable in regions that are geometrically more uniform.

\begin{table*}[ht!]
\footnotesize
\centering
\begin{tabularx}{.95\linewidth}{|*{1}{@{}>{\setlength{\hsize}{1.4\hsize}}X@{}|}|*{6}{@{}>{\setlength{\hsize}{0.95\hsize}}X@{}|}|*{5}{@{}>{\setlength{\hsize}{0.95\hsize}}X@{}|}}
\hline
 & OpenCV \cite{Drost2010} & Super 4PCS \cite{Mellado2014} & PCL \cite{Rusu2009,Holz2015} & FGR \cite{Zhou2016} & CZK \cite{Choi2015} & 3DMatch \cite{Zeng2016} & FGR with CGF-32 & CZK with CGF-32\\\hline
Recall~(\%)& 5.3 & 17.8 & 44.9 & 51.1& 59.2 & 65.1 & 60.7 & \textbf{72.0}\\\hline
Precision~(\%) & 1.6 & 10.4 & 14.0 & 23.2 & 19.6 & \textbf{25.2} & 9.4 & 14.6 \\\hline
\end{tabularx}
\normalsize
\vspace{2mm}
\caption{Evaluation on the Redwood benchmark. Plugging our learned descriptor into a pre-existing registration pipeline (CZK) yields the highest recall reported on the benchmark to date, with no training or fine-tuning on this dataset.}
\label{tab:matching}
\end{table*}

\subsection{Geometric registration}
\label{sec:registration}

We now evaluate the utility of CGF in geometric registration. For this purpose, we use Fast Global Registration (FGR)~\cite{Zhou2016}, a state-of-the-art global registration algorithm that relies on feature matching. Since feature matching is the computational bottleneck of the algorithm, using a compact feature is important. The authors' implementation of FGR uses FPFH~\cite{Zhou2016}. We use the published FGR pipeline as the baseline. To evaluate the utility of the learned CGF descriptor, we simply replace FPFH by CGF in the FGR pipeline.


To evaluate geometric registration accuracy, we follow the evaluation protocol of Choi et al.~\cite{Choi2015}, which was also used by Zhou et al.~\cite{Zhou2016} and Zeng et al.~\cite{Zeng2016} .

\mypara{Laser scans.}
We compare the accuracy of FGR when FPFH is used to the accuracy of FGR when CGF-32 is used. On the laser scan test set, FGR with FPFH correctly aligns $82.96\%$ of the pairs while FGR with CGF-32 correctly aligns $92.27\%$. The average RMSE of FGR with FPFH is $13.8\%$ of the diameter, while the average RMSE of FGR with CGF-32 is $9.2\%$ of the diameter.

\mypara{SceneNN.}
On the SceneNN test set, FGR with FPFH correctly aligns $88.54\%$ of the fragment pairs, while FGR with CGF-32 correctly aligns $91.19\%$. The average RMSE of FGR with FPFH is 14.86 cm, while the average RMSE of FGR with CGF-32 is 11.83 cm.

If we focus on the correctly aligned pairs and evaluate the average RMSE only across those, FGR with FPFH yields an RMSE of 4.68 cm and FGR with CGF-32 yields an average RMSE of 4.07 cm. This in effect evaluates the tightness of the alignment produced by global registration. \mbox{CGF-32} provides more precise correspondence pairs, which yield tighter alignment.

\mypara{Cross-dataset generalization: Redwood benchmark.}
We now evaluate on the global registration benchmark of Choi et al.~\cite{Choi2015}. This benchmark has four datasets, each containing tens of scene fragments. Geometric registration is performed on every pair of fragments, with no initialization. For this experiment, we use the feature embedding that was trained on the SceneNN dataset. We did not retrain or fine-tune the descriptor in any way. This demonstrates the learned descriptor's ability to generalize to new datasets, as well as its ability to serve as a drop-in replacement in preexisting pipelines that depend upon discriminative geometric features.

The results are reported in Table~\ref{tab:matching}. We report all the baselines from the evaluation conducted on this dataset by Zhou et al.~\cite{Zhou2016}. We plug CGF-32 into FGR~\cite{Zhou2016} and CZK~\cite{Choi2015}, the existing implementations of which use the FPFH feature. This yields the corresponding ``FGR with CGF-32'' and ``CZK with CGF-32'' conditions.

CGF improves the recall of each method by more than 9 percentage points. With \mbox{CGF-32}, the CZK pipeline achieves a recall of 72\%, by far the highest reported on the benchmark. Note that this is 6.9 percentage points higher than the contemporaneous results of Zeng et al.~\cite{Zeng2016}.


Choi et al.~\cite{Choi2015} defined two evaluation measures: recall and precision. Recall is the primary measure. The importance of recall is driven by two factors. First, the maximal level of precision that can be achieved by pairwise registration methods is low due to symmetric structures and other sources of geometric aliasing. Second, there are known ways to raise precision. Given a set of pairwise alignments, robust optimization of all fragments can prune false positives, retaining a given level of recall but increasing precision dramatically~\cite{Choi2015}.

The effect of robust optimization is demonstrated in Table~\ref{tab:pruning}. Given pairwise alignments produced by CZK with CGF-32 features, robust optimization removes false positives and yields a set of pairwise alignments with $71.1\%$ recall and $95.1\%$ precision. The accuracy of this final result is limited not by the precision of the input set of pairwise alignments~-- as the results demonstrate, the overall pipeline is robust to low precision~-- but by the level of recall. Similar precision can be achieved by applying the framework of Choi et al.~\cite{Choi2015} to any of the prior works in Table~\ref{tab:matching}.

\begin{table}[h!]
\begin{center}
\footnotesize
\begin{tabularx}{\linewidth}{|*{1}{@{}>{\setlength{\hsize}{1.2\hsize}}X@{}|}|*{4}{@{}>{\setlength{\hsize}{0.95\hsize}}X@{}|}}
\hline
\multirow{2}{*}{ } & \multicolumn{2}{c|}{Before pruning} & \multicolumn{2}{c|}{After pruning}\\\cline{2-5}
& FGR with CGF-32 & CZK with CGF-32 & FGR with CGF-32 & CZK with CGF-32 \\\hline
Recall (\%) & 60.7{} & 72.0{} & 60.7{} & 71.1{}\\\hline
Precision (\%) & 9.4{} & 14.6{} & 86.8{} & 95.1{}\\\hline
\end{tabularx}
\normalsize
\end{center}
\caption{After post-processing with robust global optimization~\cite{Choi2015}, CZK with CGF-32 achieves 71.1\% recall and 95.1\% precision on the Redwood benchmark.}
\label{tab:pruning}
\end{table}



\section{Conclusion}
\label{doc:conclusion}
We presented an approach to obtaining discriminative features for local geometry in unstructured point clouds. We have shown that state-of-the-art accuracy can be achieved with a low-dimensional feature space. The learned descriptor is both more precise and more compact than hand-crafted features. Due to its Euclidean structure, the learned descriptor can be used as a drop-in replacement for existing features in robotics, 3D vision, and computer graphics applications.
We expect future work to further improve precision, compactness, and robustness, possibly using new approaches to optimizing feature embeddings~\cite{Ustinova2016}.

\balance

{\small
\bibliographystyle{ieee}
\bibliography{paper_learningfeatures}

\begin{thebibliography}{10}\itemsep=-1pt

\bibitem{Aubry2011}
M.~Aubry, U.~Schlickewei, and D.~Cremers.
\newblock The wave kernel signature: A quantum mechanical approach to shape
  analysis.
\newblock In {\em ICCV Workshops}, 2011.

\bibitem{Babenko2014}
A.~Babenko, A.~Slesarev, A.~Chigorin, and V.~S. Lempitsky.
\newblock Neural codes for image retrieval.
\newblock In {\em ECCV}, 2014.

\bibitem{Balntas2016}
V.~Balntas, E.~Johns, L.~Tang, and K.~Mikolajczyk.
\newblock {PN-Net}: Conjoined triple deep network for learning local image
  descriptors.
\newblock {\em arXiv:1601.05030}, 2016.

\bibitem{Besl1992}
P.~J. Besl and N.~D. McKay.
\newblock A method for registration of {3-D} shapes.
\newblock {\em PAMI}, 14(2), 1992.

\bibitem{Boscaini2016}
D.~Boscaini, J.~Masci, E.~Rodol{\`{a}}, and M.~M. Bronstein.
\newblock Learning shape correspondence with anisotropic convolutional neural
  networks.
\newblock In {\em NIPS}, 2016.

\bibitem{Chechik2010}
G.~Chechik, V.~Sharma, U.~Shalit, and S.~Bengio.
\newblock Large scale online learning of image similarity through ranking.
\newblock {\em JMLR}, 11, 2010.

\bibitem{Choi2015}
S.~Choi, Q.-Y. Zhou, and V.~Koltun.
\newblock Robust reconstruction of indoor scenes.
\newblock In {\em CVPR}, 2015.

\bibitem{Cosmo2016}
L.~Cosmo, E.~Rodol{\`{a}}, J.~Masci, A.~Torsello, and M.~M. Bronstein.
\newblock Matching deformable objects in clutter.
\newblock In {\em 3DV}, 2016.

\bibitem{Drost2010}
B.~Drost, M.~Ulrich, N.~Navab, and S.~Ilic.
\newblock Model globally, match locally: Efficient and robust {3D} object
  recognition.
\newblock In {\em CVPR}, 2010.

\bibitem{Elseberg2012}
J.~Elseberg, S.~Magnenat, R.~Siegwart, and A.~N{\"u}chter.
\newblock Comparison of nearest-neighbor-search strategies and implementations
  for efficient shape registration.
\newblock {\em Journal of Software Engineering for Robotics}, 3(1), 2012.

\bibitem{Frome2004}
A.~Frome, D.~Huber, R.~Kolluri, T.~B{\"u}low, and J.~Malik.
\newblock Recognizing objects in range data using regional point descriptors.
\newblock In {\em ECCV}, 2004.

\bibitem{Guo2016}
Y.~Guo, M.~Bennamoun, F.~A. Sohel, M.~Lu, J.~Wan, and N.~M. Kwok.
\newblock A comprehensive performance evaluation of {3D} local feature
  descriptors.
\newblock {\em IJCV}, 116(1), 2016.

\bibitem{Guo2013}
Y.~Guo, F.~A. Sohel, M.~Bennamoun, M.~Lu, and J.~Wan.
\newblock Rotational projection statistics for {3D} local surface description
  and object recognition.
\newblock {\em IJCV}, 105(1), 2013.

\bibitem{Han2015}
X.~Han, T.~Leung, Y.~Jia, R.~Sukthankar, and A.~C. Berg.
\newblock {MatchNet}: Unifying feature and metric learning for patch-based
  matching.
\newblock In {\em CVPR}, 2015.

\bibitem{He2016}
K.~He, X.~Zhang, S.~Ren, and J.~Sun.
\newblock Deep residual learning for image recognition.
\newblock In {\em CVPR}, 2016.

\bibitem{Holz2015}
D.~Holz, A.~E. Ichim, F.~Tombari, R.~B. Rusu, and S.~Behnke.
\newblock Registration with the point cloud library: A modular framework for
  aligning in {3-D}.
\newblock {\em IEEE Robotics and Automation Magazine}, 22(4), 2015.

\bibitem{Scenenn2016}
B.-S. Hua, Q.-H. Pham, D.~T. Nguyen, M.-K. Tran, L.-F. Yu, and S.-K. Yeung.
\newblock {SceneNN}: A scene meshes dataset with annotations.
\newblock In {\em 3DV}, 2016.

\bibitem{Itoh1995}
T.~Itoh and K.~Koyamada.
\newblock Automatic isosurface propagation using an extrema graph and sorted
  boundary cell lists.
\newblock {\em TVCG}, 1(4), 1995.

\bibitem{JohnsonHebert1999}
A.~E. Johnson and M.~Hebert.
\newblock Using spin images for efficient object recognition in cluttered {3D}
  scenes.
\newblock {\em PAMI}, 21(5), 1999.

\bibitem{KingmaBa2015}
D.~P. Kingma and J.~Ba.
\newblock Adam: {A} method for stochastic optimization.
\newblock In {\em ICLR}, 2015.

\bibitem{Kolluri2004}
R.~K. Kolluri, J.~R. Shewchuk, and J.~F. O'Brien.
\newblock Spectral surface reconstruction from noisy point clouds.
\newblock In {\em Symposium on Geometry Processing}, 2004.

\bibitem{MaturanaScherer2015}
D.~Maturana and S.~Scherer.
\newblock {VoxNet}: {A} {3D} convolutional neural network for real-time object
  recognition.
\newblock In {\em IROS}, 2015.

\bibitem{Mellado2014}
N.~Mellado, D.~Aiger, and N.~J. Mitra.
\newblock {Super 4PCS}: Fast global pointcloud registration via smart indexing.
\newblock {\em Computer Graphics Forum}, 33(5), 2014.

\bibitem{flann2014}
M.~Muja and D.~G. Lowe.
\newblock Scalable nearest neighbor algorithms for high dimensional data.
\newblock {\em PAMI}, 36, 2014.

\bibitem{Rusu2009}
R.~B. Rusu, N.~Blodow, and M.~Beetz.
\newblock Fast point feature histograms {(FPFH)} for {3D} registration.
\newblock In {\em ICRA}, 2009.

\bibitem{Rusu2008-IROS}
R.~B. Rusu, N.~Blodow, Z.~C. Marton, and M.~Beetz.
\newblock Aligning point cloud views using persistent feature histograms.
\newblock In {\em IROS}, 2008.

\bibitem{Salti2014}
S.~Salti, F.~Tombari, and L.~di~Stefano.
\newblock {SHOT}: Unique signatures of histograms for surface and texture
  description.
\newblock {\em Computer Vision and Image Understanding}, 125, 2014.

\bibitem{Schroff2015}
F.~Schroff, D.~Kalenichenko, and J.~Philbin.
\newblock {FaceNet}: {A} unified embedding for face recognition and clustering.
\newblock In {\em CVPR}, 2015.

\bibitem{SchultzJoachims2003}
M.~Schultz and T.~Joachims.
\newblock Learning a distance metric from relative comparisons.
\newblock In {\em NIPS}, 2003.

\bibitem{Simo-Serra2015}
E.~Simo{-}Serra, E.~Trulls, L.~Ferraz, I.~Kokkinos, P.~Fua, and
  F.~Moreno{-}Noguer.
\newblock Discriminative learning of deep convolutional feature point
  descriptors.
\newblock In {\em ICCV}, 2015.

\bibitem{Su2015}
H.~Su, S.~Maji, E.~Kalogerakis, and E.~G. Learned{-}Miller.
\newblock Multi-view convolutional neural networks for {3D} shape recognition.
\newblock In {\em ICCV}, 2015.

\bibitem{Sun2009}
J.~Sun, M.~Ovsjanikov, and L.~J. Guibas.
\newblock A concise and provably informative multi-scale signature based on
  heat diffusion.
\newblock {\em Computer Graphics Forum}, 28(5), 2009.

\bibitem{Tombari2010b}
F.~Tombari, S.~Salti, and L.~Di~Stefano.
\newblock Unique shape context for {3D} data description.
\newblock In {\em ACM Workshop on 3D Object Retrieval}, 2010.

\bibitem{Ustinova2016}
E.~Ustinova and V.~Lempitsky.
\newblock Learning deep embeddings with histogram loss.
\newblock In {\em NIPS}, 2016.

\bibitem{Wei2016}
L.~Wei, Q.~Huang, D.~Ceylan, E.~Vouga, and H.~Li.
\newblock Dense human body correspondences using convolutional networks.
\newblock In {\em CVPR}, 2016.

\bibitem{Weinberger2009}
K.~Q. Weinberger and L.~K. Saul.
\newblock Distance metric learning for large margin nearest neighbor
  classification.
\newblock {\em JMLR}, 10, 2009.

\bibitem{Wu2015}
Z.~Wu, S.~Song, A.~Khosla, F.~Yu, L.~Zhang, X.~Tang, and J.~Xiao.
\newblock {3D ShapeNets}: {A} deep representation for volumetric shapes.
\newblock In {\em CVPR}, 2015.

\bibitem{Yi2016}
K.~M. Yi, E.~Trulls, V.~Lepetit, and P.~Fua.
\newblock {LIFT}: Learned invariant feature transform.
\newblock In {\em ECCV}, 2016.

\bibitem{ZagoruykoKomodakis2015}
S.~Zagoruyko and N.~Komodakis.
\newblock Learning to compare image patches via convolutional neural networks.
\newblock In {\em CVPR}, 2015.

\bibitem{Zbontar2016}
J.~{\v{Z}}bontar and Y.~LeCun.
\newblock Stereo matching by training a convolutional neural network to compare
  image patches.
\newblock {\em JMLR}, 17, 2016.

\bibitem{Zeng2016}
A.~Zeng, S.~Song, M.~Nie{\ss}ner, M.~Fisher, J.~Xiao, and T.~Funkhouser.
\newblock {3DMatch}: Learning the matching of local {3D} geometry in range
  scans.
\newblock In {\em CVPR}, 2017.

\bibitem{Zhou2016}
Q.-Y. Zhou, J.~Park, and V.~Koltun.
\newblock Fast global registration.
\newblock In {\em ECCV}, 2016.

\end{thebibliography}
}

\end{document}